\documentclass[10pt,twocolumn,letterpaper]{article}
\usepackage{iccv}
\usepackage{times}
\usepackage{epsfig}
\usepackage{graphicx}
\usepackage{amsmath}
\usepackage{amssymb}
\usepackage{diagbox}
\usepackage{adjustbox}
\usepackage{booktabs}
\usepackage{makecell}
\usepackage{commath}
\usepackage[dvipsnames]{xcolor}
\usepackage{multirow}
\usepackage{standalone}
\usepackage{subfiles}

\newcommand{\ourmodel}{RobustSwap\xspace}

\usepackage[pagebackref=true,breaklinks=true,letterpaper=true,colorlinks,bookmarks=false]{hyperref}

\iccvfinalcopy 

\def\iccvPaperID{6821} 

\begin{document}


\title{RobustSwap: A Simple yet Robust Face Swapping Model \\ against Attribute Leakage}

\author{
Jaeseong Lee\textsuperscript{*}\textsuperscript{\rm 1}\quad\quad 
Taewoo Kim\textsuperscript{*}\textsuperscript{\rm 2}\quad\quad \\
Sunghyun Park\textsuperscript{\rm 1}\quad\quad
Younggun Lee\textsuperscript{\rm 2}\quad\quad 
Jaegul Choo\textsuperscript{\rm 1}\\ \\
\textsuperscript{\rm 1} KAIST\quad\quad
\textsuperscript{\rm 2} Neosapience\quad\quad\\
{\tt\small \{wintermad, psh01087, jchoo\}@kaist.ac.kr,} \\ 
{\tt\small \{taewoo, yg\}@neosapience.com} \\ \\
{\tt\large \href{https://robustswap.github.io/}{https://robustswap.github.io/}}\\
}
\newcommand\blfootnote[1]{
  \begingroup
  \renewcommand\thefootnote{}\footnote{#1}
  \addtocounter{footnote}{-1}
  \endgroup
}

\blfootnote{\textsuperscript{*} These authors contributed equally.}

\ificcvfinal\thispagestyle{empty}\fi

\begin{figure}
\twocolumn[{
\renewcommand\twocolumn[1][]{#1}
    \centering 
    \maketitle
    \vspace{-0.7cm}
    \includegraphics[width=\linewidth]{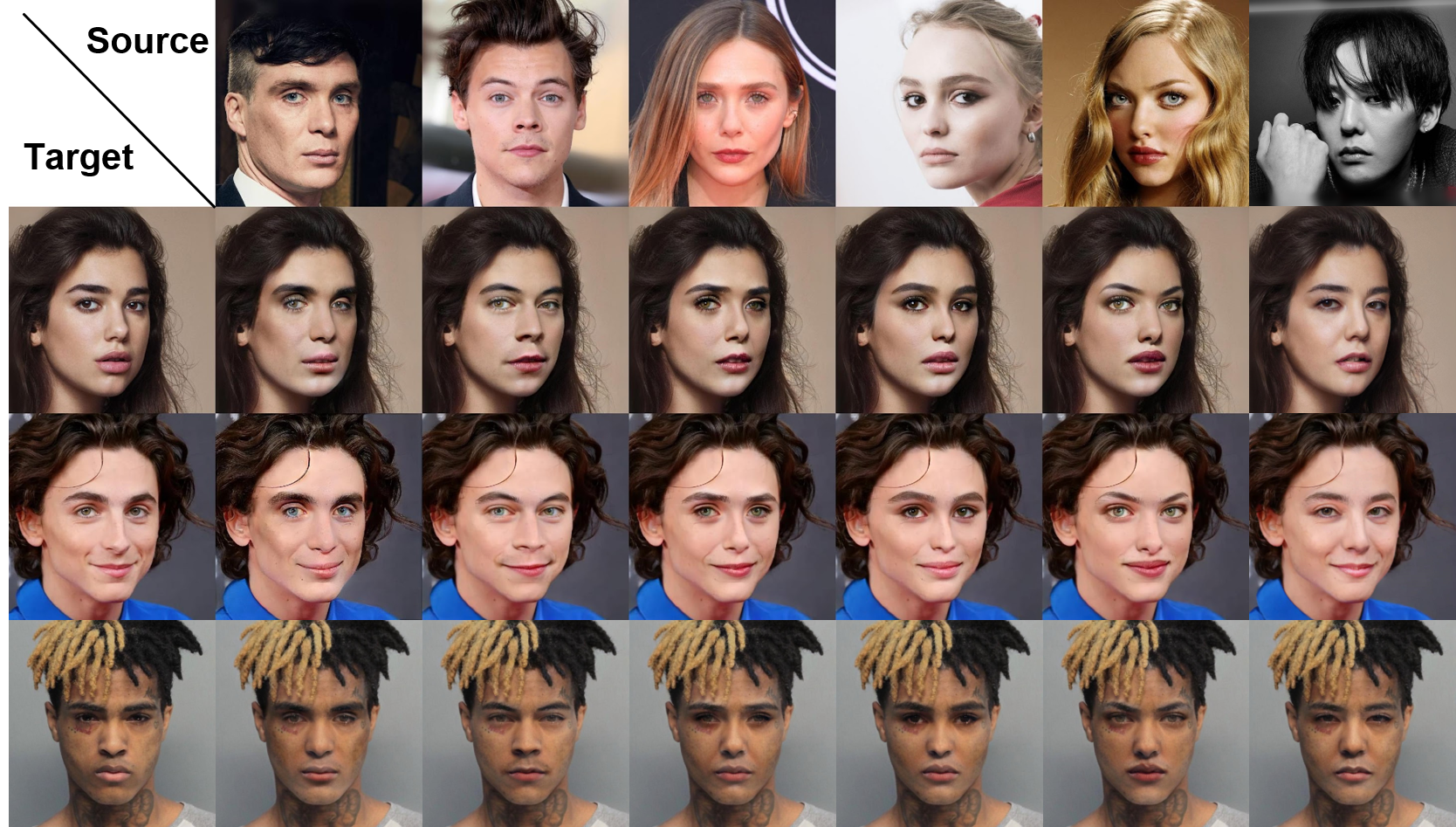}
    \caption{\textbf{Face swapping results of \ourmodel.} Our model robustly synthesizes the swapped image from the in-the-wild source and target images, while preserving the target attributes and reflecting source identity. With our model, the target attributes (\textit{e.g.}, eye gazing, hair, background, expression, and head pose) are preserved successfully regardless of source identity images.}
    \vspace{0.1cm}
\label{fig:teaser}
}]
\end{figure}


\begin{abstract}

Face swapping aims at injecting a source image's identity (i.e., facial features) into a target image, while strictly preserving the target's attributes, which are irrelevant to identity.
However, we observed that previous approaches still suffer from source attribute leakage, where the source image's attributes interfere with the target image's.
In this paper, we analyze the latent space of StyleGAN and find the adequate combination of the latents geared for face swapping task.
Based on the findings, we develop a simple yet robust face swapping model, \textbf{\ourmodel}, which is resistant to the potential source attribute leakage.
Moreover, we exploit the coordination of 3DMM's implicit and explicit information as a guidance to incorporate the structure of the source image and the precise pose of the target image.
Despite our method solely utilizing an image dataset without identity labels for training, our model has the capability to generate high-fidelity and temporally consistent videos.
Through extensive qualitative and quantitative evaluations, we demonstrate that our method shows significant improvements compared with the previous face swapping models in synthesizing both images and videos.

\end{abstract}

\vspace{-0.5cm}
\section{Introduction}
\label{sec:intro}

Face swapping has become a prominent task with various applications such as digital resurrection, virtual human avatars, and movie films.
The goal of face swapping is to inject a source's identity (\textit{e.g.}, eyes, nose, lips, and eyebrows) into a target, while strictly preserving the target's attributes (\textit{e.g.}, hair, background, light condition, expression, head pose, and eye gazing), which are irrelevant to identity.
Due to the notorious intractability of protecting the target person's attributes against potential interference by the source person's attributes, previous research has endeavored to overcome this challenge.
Two primary categories of face swapping approaches exist.

In one approach to face swapping, the reconstruction loss between the swapped and target images is employed when the source and target images share the same identity.~\cite{simswap,hififace,faceshifter,styleswap,fslsd,uniswap}.
However, applying the reconstruction loss in certain scenarios necessitates the use of identity-labeled image datasets~\cite{vggface1,vggface2} or video datasets~\cite{voxceleb1,voxceleb2}.
Unfortunately, it is challenging to obtain high-quality images with identity labels, hence limiting the applicability of these methods.
Moreover, these methods require careful hyperparameter tuning to determine the appropriate ratio between the same and cross-identity images.

To synthesize high-resolution images, the other approaches utilize a pre-trained StyleGAN model as a strong prior with layer-wise information injection~\cite{megapixel,mfim,fslsd}.
Despite the power of the pre-trained StyleGAN, MegaFS~\cite{megapixel} and FSLSD~\cite{fslsd} often fail to preserve the target person's attributes.
This issue stems from utilizing solely $\mathcal{W+}$ space for assembling the latent codes in StyleGAN from the source and target images.
To preserve the target person's attributes, MFIM~\cite{mfim} replaces the spatial noise maps of StyleGAN with the spatially-dimensioned feature maps of the target image. 
However, we found that their empirically designed architecture still induces low-fidelity results that are affected by the source person's attributes, such as the source person's hair and eyeglasses.

\begin{figure}[t!]
    \centering 
    \includegraphics[width=\linewidth]{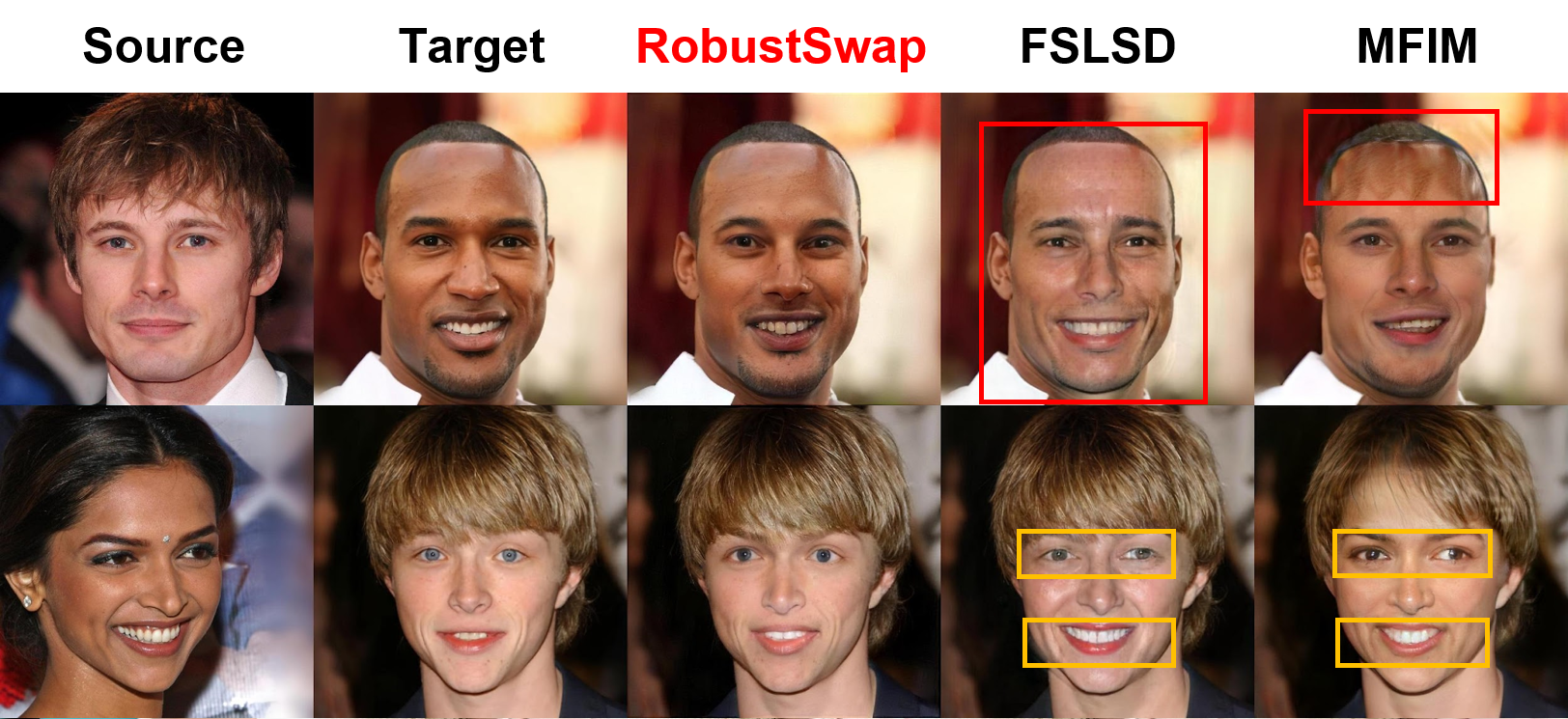}
    \caption{\textbf{Examples of source attribute leakage} and our improved results; In the first row, FSLSD~\cite{fslsd} often fails to preserve the skin color and lighting condition of the target image. MFIM~\cite{mfim} brings hairstyle from the source; In the second row, FSLSD~\cite{fslsd} and MFIM~\cite{mfim} hardly preserve the target image's pose such as eye gazing and expression. Besides, our result has no artifacts like those.
    \textcolor{Dandelion}{Yellow} boxes indicate the \textit{appearance leakage}.
    \textcolor{red}{Red} boxes indicate the \textit{pose leakage}.}
    \vspace{-0.5cm}
    \label{fig:leakage}
\end{figure}

Although previous studies struggle to balance the information between the source and target images, they are still vulnerable to \textbf{source attribute leakage} problem, defined as \textit{source person's identity irrelevant information leaking to the target person's image}.
For example, as shown in the first row of Fig.~\ref{fig:leakage}, the existing face swapping methods often bring the source image's appearance to the target image, such as hair and skin color, which is defined as \textit{appearance leakage}. 
In the second row of Fig.~\ref{fig:leakage}, the source's pose (\textit{e.g.}, head pose, expression, and eye gazing) interferes with the target's pose, which is defined as \textit{pose leakage}.

To solve these \textbf{source attribute leakages}, we thoughtfully design a simple yet robust face swapping model called \textbf{\ourmodel}, which employs a pre-trained StyleGAN~\cite{sg2}.
Behind our model, we explore StyleGAN's latent space $\mathcal{F}/\mathcal{W+}$ to find the promising combination of latents in the subspaces for preventing \textbf{source attribute leakage}.
In specific, we investigate the suitable latents by assessing the extent to which the target's pose can be changed at each combination of latents in subspaces.
Armed with the investigation, we elaborately design a face swapping model, which is robust to preserving the target image's attributes, while effectively reflecting the source image's identity.
  
To impose the detailed face shape information of the source image, our model takes the source's shape parameter of 3D Morphable Model (3DMM)~\cite{flame,bfm,ls3dmm} as the input.
In addition to inject the shape parameters into the model, we introduce a novel partial landmark loss, which is effective to retain the head pose and expression of the target image, while injecting the inner facial geometry of the source image.
Thanks to our well-designed simple architecture and the coordination of the 3DMM information, \textbf{\ourmodel} is secured from the \textbf{source attribute leakage} and injects the more abundant identity information.
Moreover, \textbf{\ourmodel} is built on megapixels (\textit{e.g.}, 1024 $\times$ 1024), which is practical and applicable in various applications.

In summary, our contributions are three-fold.
\begin{itemize}
    \item Based on the analysis of StyleGAN latent space, we introduce \textbf{\ourmodel}, preserving target attributes while preventing the \textbf{source attribute leakage}. 
    \item For casting detailed source identity information and precise target's pose, we propose a shape-guided identity condition and a partial landmark loss with 3DMM. 
    \item Extensive experiments demonstrate that \textbf{\ourmodel} outperforms previous approaches quantitatively and qualitatively. 
    Moreover, \textbf{\ourmodel} can produce high-quality videos without training on video datasets. 
\end{itemize}


\section{Related Work}

\noindent\textbf{Face Swapping.}
There are numerous face swapping methods employing identity-labeled datasets. 
FaceShifter~\cite{faceshifter} designs its occlusion-aware architecture with two stages.
SimSwap~\cite{simswap} devises a robust method via weak feature-matching loss. 
InfoSwap~\cite{infoswap} utilizes the information-bottleneck principle for disentangling identity-attribute information. 
HifiFace~\cite{hififace} firstly exploits 3DMM's semantic information in face swapping.
StyleSwap~\cite{styleswap} uses simple modification of StyleGAN with the identity-labeled datasets for training. 
However, the usability of these methods is restricted due to the challenge of obtaining high-quality images with identity labels or video datasets.
Moreover, they necessitate careful hyperparameter tuning to determine the appropriate ratio between the same and cross-identity images. 
In contrast, \textbf{\ourmodel} is trained on a high-quality image dataset~\cite{sg1}, eliminating the need for searching for the appropriate ratio.
To generate high-resolution images, recent face swapping approaches, such as MegaFS~\cite{megapixel}, FSLSD~\cite{fslsd}, and MFIM~\cite{mfim}, employ a pre-trained StyleGAN~\cite{sg2} as a strong prior.
However, we discover that those methods based on the pre-trained StyleGAN fail to prevent \textbf{source attribute leakage} problem.
Different from previous studies, we conduct a depth experiment to seek the face swapping adaptive latent space of StyleGAN and appropriate architecture.

\noindent\textbf{StyleGAN's Latent Space.} 
StyleGANs~\cite{sg1, sg2, sg3} have shown remarkable success in generating realistic images. 
Following the success of the StyleGANs, the latent space of StyleGAN has been the subject of recent studies, with exploring various aspects of its properties and dynamics.
In the previous StyleGAN inversion studies~\cite{image2stylegan, image2stylegan++, psp, e4e}, they expand the $\mathcal{W}$ space to $\mathcal{W+}$ to amplify the StyleGAN's representation capacity.
Moreover, a previous study~\cite{oorinversion} proposes a method that maps images to an alternative latent space $\mathcal{F}/\mathcal{W+}$ in StyleGAN, which allows for more accurate reconstruction and semantic editing of out-of-range images with geometric transformations and local variations.
Also, numerous recent work~\cite{barbershop, styleyourhair, wang2022high, styleheat} utilize the latent feature map space $\mathcal{F}$, which is spatial-aware, to keep spatial information to be maintained while manipulating other traits.
They demonstrate the potential of the latent feature map space $\mathcal{F}$ in StyleGAN for a variety of image manipulation tasks.
Inspired by these findings and applications, we investigate the suitability of $\mathcal{F}/\mathcal{W+}$ for face swapping task and find which combination of the subspaces is proper in respective of face swapping.
To achieve this goal, we conduct a detailed experiment to explore the $\mathcal{F}/\mathcal{W+}$ space of StyleGAN, and analyze the subspaces to design a robust face swapping model.

\begin{figure}[t!]
    \centering 
    \includegraphics[width=\linewidth]{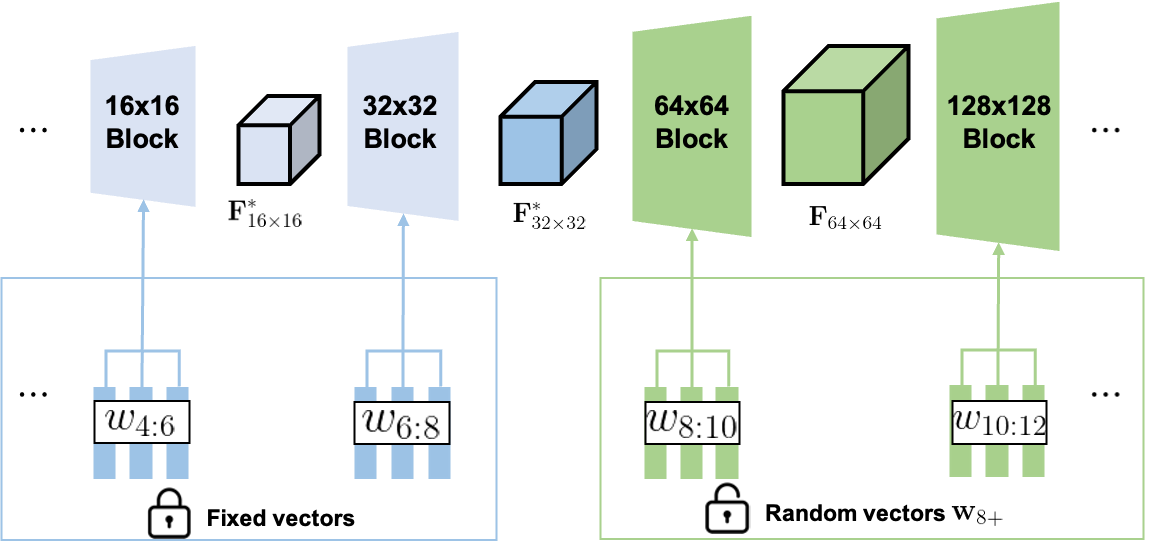}
    \vspace{-0.2cm}
    \caption{\textbf{Analysis process of $\mathcal{F}/\mathcal{W+}$} with pre-trained StyleGAN; we generate \textit{random sampled} images with a fixed feature map $\mathbf{F}^{*}_{h \times w}$ and $\mathbf{w_{m+}}$, and an \textit{anchor} image is obtained from $\mathbf{w_{1+}}$.}
    \vspace{-0.65cm}
    \label{fig:short}
\end{figure} 

\noindent\textbf{3D Morphable Models.} 
A 3D morphable face model (3DMM)~\cite{bfm,flame,ls3dmm} is a strong representation for modeling human faces, including head pose, shape, and expression.
The 3DMM's shape is transformed into a PCA-based vector space, which can fit the human faces into the vector space.
Consequently, their corresponding encoders~\cite{deep3drecon,deca,ringnet} have came out to alleviate the time-consuming optimization. 
We utilize the 3DMM's shape parameter from the state-of-the-art~\cite{deca} 3DMM encoder, and corresponding decoder~\cite{flame} for our partial landmark loss.



\section{Method}

Given a source identity image $I_{src} \in \mathbb{R}^{H \times W \times 3}$ and target attribute image $I_{tgt} \in \mathbb{R}^{H \times W \times 3}$, our goal is to inject the identity of $I_{src}$ to $I_{tgt}$, while preserving the attribute of $I_{tgt}$ to synthesize the swapped image $\hat{I}$.
$H$ and $W$ indicate the height and width of the image, respectively. 
We explore latent subspaces $\mathcal{F/W+}$ of StyleGAN~\cite{sg2} to analyze the degree of variation in aspects of identity and attributes (Section~\ref{one}).
Through the analysis, we find the appropriate combination of latents, which can preserve the attribute of $I_{tgt}$, while reflecting the identity of $I_{src}$.
We introduce our face swapping model, \textbf{\ourmodel}, which is robust to the \textbf{source attribute leakage} (Section~\ref{two}).
Last but not least, we describe the objective functions for our method, including a novel partial landmark loss, which coordinates with 3DMM's implicit shape information (Section~\ref{three}).

\subsection{Exploring StyleGAN for Face Swapping.}\label{one}

In this section, we analyze the latent space of StyleGAN~\cite{sg2} from the perspective of developing the face swapping model.
Then, we justify the proper combination of latents for a \textbf{source attribute leakage}-free model.

\noindent\textbf{Revisiting Latent Space of StyleGAN.}
StyleGAN is a generative model that produces a high-resolution image $\hat{I} \in \mathbb{R}^{H \times W \times 3}$ using $n$ identical vector $\textbf{w} \in \mathcal{W} \subsetneq \mathbb{R}^{1 \times 512}$.
Recent work on GAN inversion~\cite{oorinversion,wang2022high} split the latent space of StyleGAN into two subspaces: the latent vector space $\mathcal{W+}$ and latent feature map space $\mathcal{F}$.
The extended latent vectors $\{w_1,w_2, \cdots, w_n\} \in \mathcal{W+} \subsetneq \mathbb{R}^{n \times 512}$, used for the different StyleGAN layers, allow StyleGAN to represent the diverse images and fine-grained control over the generated images. 
However, due to the deficient spatial information in $\mathcal{W+}$, it is difficult to reconstruct the structural details of images.
To address this problem, latent spatial feature map $\mathbf{F}_{h \times w} \in \mathbb{R}^{h \times w \times c}$, which is in $\mathcal{F}$, is used to represent the details of spatial information, where $h$, $w$, and $c$ are height, width, and channel dimension of the feature map, respectively.
With $\mathcal{W+}$ and $\mathcal{F}$, StyleGAN is reformulated as:
\begin{equation}
    \hat{I} = G(\mathbf{F}_{h \times w},\textbf{w}_{m+}),
\end{equation}
where $\textbf{w}_{m+}$ = $\{w_{m}, w_{m+1}, \cdots, w_{n}\}$$ \subset \mathcal{W+}.$
Note that the $\mathbf{F}_{h \times w}$ and $\textbf{w}_{m+}$ are complementary to each other.~\footnote{The $h \times w$ block maps ($\{{w_{m-2}, w_{m-1}}\}$, $\mathbf{F}_{h/2 \times w/2}$) to $\mathbf{F}_{h \times w}$. Please refer the Fig.~\ref{fig:short}. and Fig.~\ref{fig:model} (B)} 

\begin{figure}[t!]
    \centering 
    \includegraphics[width=.85\linewidth]{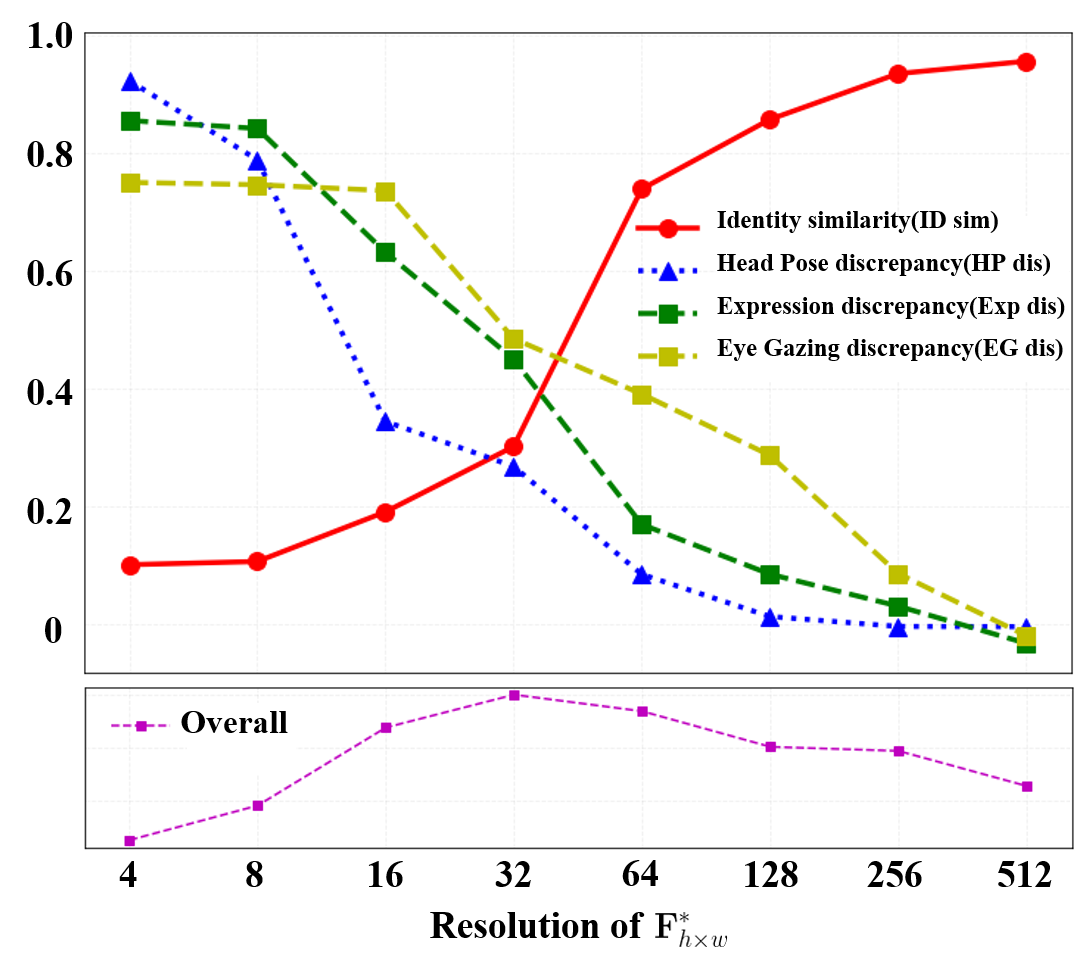}
    \vspace{-0.3cm}
    \caption{\textbf{Quantitative analysis} between \textit{anchor} and \textit{random sampled}. 
    The larger $\mathbf{F}_{h \times w}^{*}$ results in improved preservation of expression, head pose, and eye gazing, while the identity undergoes less change. 
    Overall score is calculated by (ID sim)$^{3} *$ (HP dis) $*$ (Exp dis) $*$ (EG dis) which is standardized. Details for each score metric are described in the supplementary materials.}
    \vspace{-0.5cm}
    \label{fig:graph}
\end{figure}

Motivated by the advantages of $\mathcal{F}/\mathcal{W+}$, the following question arises: Is it appropriate to map the target spatial attribute to $\mathcal{F}$ while injecting source identity via $\mathcal{W+}$?
However, there is a lack of studies analyzing the suitability of the StyleGAN latent space for the face swapping.
Thus, we explore the latent space $\mathcal{F}/\mathcal{W+}$; the combination of $(\mathbf{F}_{h \times w},\textbf{w}_{m+})$ for building a face swapping model.

\noindent\textbf{Analysis on $\mathcal{F}/\mathcal{W+}$ for Face Swapping.}
We study the profitable combination of $\mathbf{F}_{h \times w}$ containing spatial attributes of the target and $\textbf{w}_{m+}$ embedded the source identity from the perspective of face swapping task.
To achieve this goal, we conduct the following experiment. 
As shown in Fig.~\ref{fig:short}, we fix the $\mathbf{F}_{h \times w}$ (corresponds to target attributes) at the certain spatial resolution denoted as $\mathbf{F}^{*}_{h \times w}$, and generate images with randomly initialized $\textbf{w}_{m+}$ (corresponds to the identity of source).
Formally, it is denoted as:
\begin{equation}
    \hat{I}_{\textbf{w}_{m+}}=G(\mathbf{F}^{*}_{h \times w}, \textbf{w}_{m+}),
\end{equation}
where $\mathbf{F}^{*}_{h \times w}$ is generated from fixed vectors $\{w_1, \cdots, w_{m-1} \}$.
Then, we examine the generated image as gradually increasing the resolution of $\mathbf{F}^{*}_{h \times w}$ from $4\times 4$ to $512 \times 512$.

Now, we analyze quantitative factors to be considered in the face swapping task to find a suitable combination.
As shown in Fig.~\ref{fig:graph}, when the resolution of feature map enlarges, identity similarity increases, and head pose, expression, and eye gazing discrepancy decrease between \textit{anchor} and \textit{random sampled} images.
We assume that the most adequate combination of $\mathbf{F}_{h \times w}$ and $\textbf{w}_{m+}$ for the robust face swapping should show low identity similarity, and head pose, expression, and eye gazing discrepancy between \textit{anchor} and \textit{random sampled}, since that combination can change identity with preserving the pose information. 

\begin{figure}[t!]
    \centering 
    \vspace{-0.4cm}
    \includegraphics[width=\linewidth]{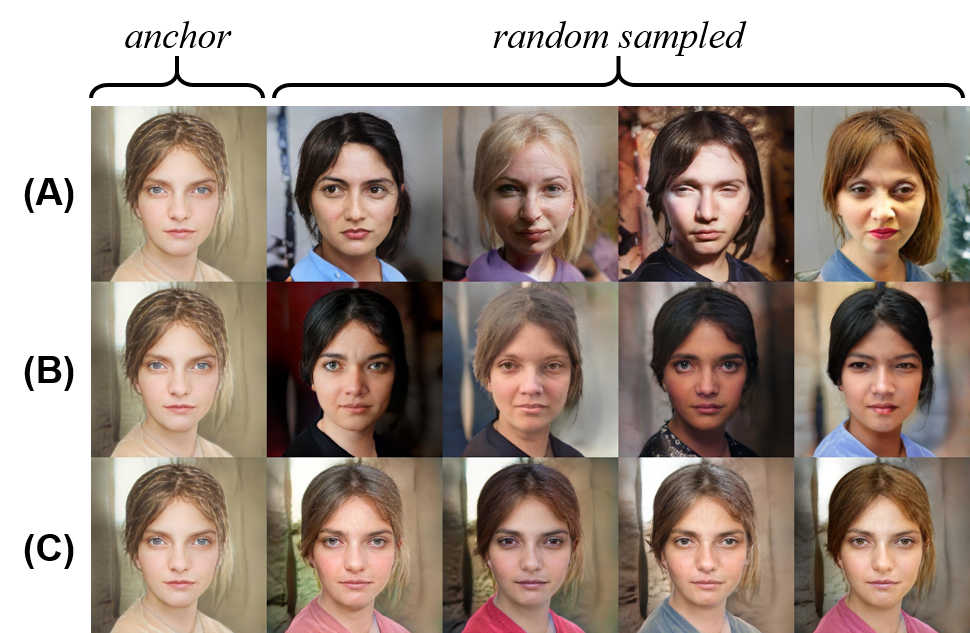}
    \caption{\textbf{Qualitative analysis.} Examples from the analysis on the latent space for face swapping. An \textit{anchor} image is obtained from the inverted vectors $\textbf{w}_{1+}$ by using GAN inversion method~\cite{psp}. \textit{Random sampled} images of (A) are generated by the fixed feature map $\mathbf{F}^{*}_{16 \times 16}$ and randomly initialized $\mathbf{w}_{6+}$. 
    (B)'s \textit{random sampled} images are produced by $\mathbf{F}^{*}_{32 \times 32}$ and $\mathbf{w}_{8+}$.
    \textit{Random sampled} images of (C) are obtained from $\mathbf{F}^{*}_{64 \times 64}$ and $\mathbf{w}_{10+}$.
    }
    \vspace{-0.6cm}
    \label{fig:toy}
\end{figure}

\begin{figure*}[t!]
    \centering 
    \includegraphics[width=\linewidth]{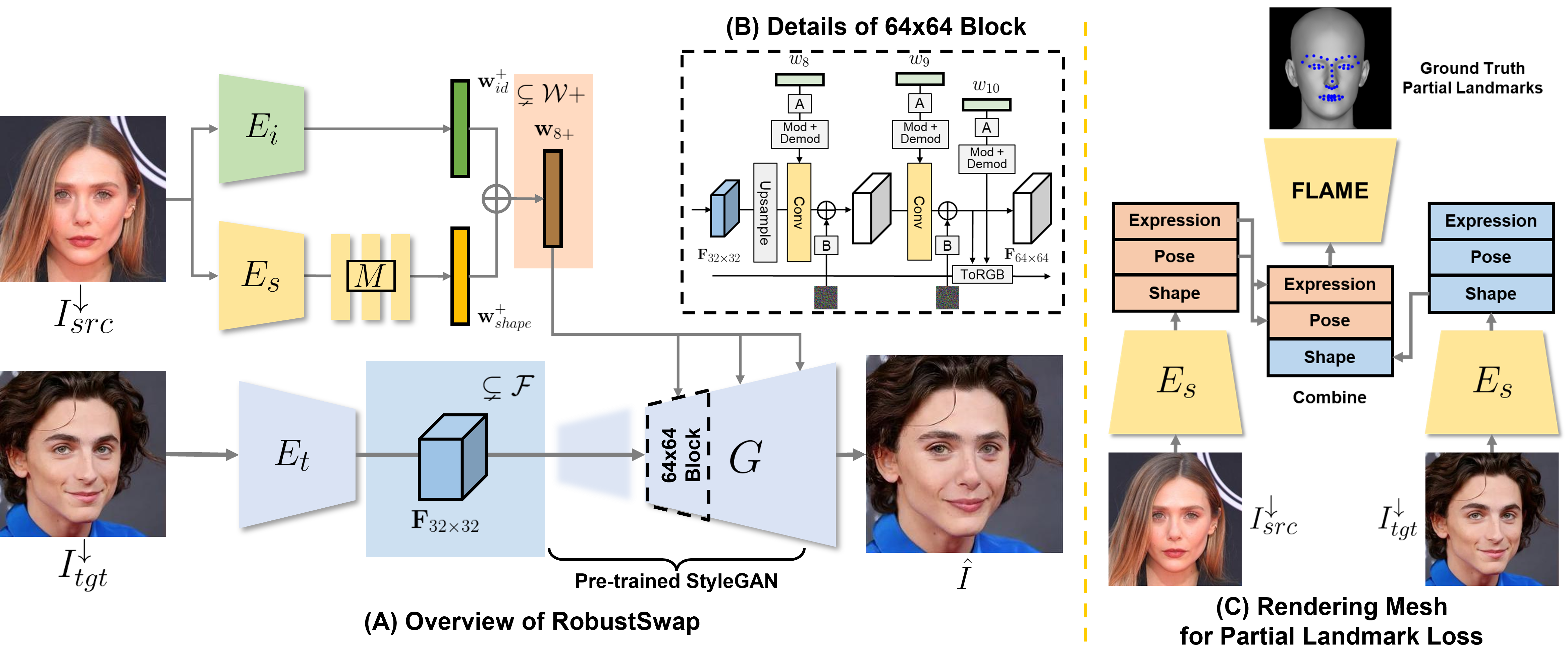}  
    \vspace{-0.7cm}
    \caption{(A) Our \textbf{\ourmodel} architecture; the blurred trapezoidal box is the area of the discarded block of StyleGAN. The target encoder $E_{t}$ encodes $I_{tgt}^{\downarrow}$ to $\mathbf{F}_{32 \times 32}$. The encoded $\textbf{w}_{8+}$ from the two source encoder $E_{i}$ and $E_{s}$ is injected to StyleGAN $G$. (B) illustrates the details of 64 x 64 Block. It produces $\mathbf{F}_{64 \times 64}$ from $\{\textbf{w}_{8}, \textbf{w}_{9}\}$ and $\mathbf{F}_{32 \times 32}$. (C) is the construction pipeline of ground-truth for partial landmark loss. More details are described in our supplementary materials. }
    \vspace{-0.5cm}
\label{fig:model}
\end{figure*}

To observe the three highest overall scored $\mathbf{F}^{*}_{h \times w}$, from 16 to 64, we visualize the \textit{anchor} and \textit{random sampled} in Fig.~\ref{fig:toy}. 
(A) varies a lot of attributes like expression and eye gazing, and (C) does not vary except for lighting conditions, skin, and background colors.
However, (B) varies inner facial parts, while the pose, eye gazing, and expression are similar to the \textit{anchor} image's.

Therefore, we select the combination of $(\mathbf{F}_{32 \times 32}, \textbf{w}_{8+})$ since it effectively preserves the pose, eye gazing, and expression while changing identity relevant features such as eyes, nose, lip, and eyebrows.
It implies that as long as we utilize $\mathbf{F}_{32 \times 32}$ and $\textbf{w}_{8+}$, the source identity is well-reflected, minimizing damage to the target attributes.

\subsection{\textbf{\ourmodel}: Simple yet Robust Architecture for Face Swapping.}\label{two}
As shown in Fig.~\ref{fig:model}, we utilize pre-trained StyleGAN without any architectural modification since $(\mathbf{F}_{32 \times 32}$, $\textbf{w}_{8+})$ preserve the target attributes while switching the identity. 

\noindent\textbf{Target Attributes Encoder.} 
Our generation pipeline starts from $\mathbf{F}_{32 \times 32}$, which is encoded from $I_{tgt}$. 
To directly map the spatial information of $I_{tgt}$ to the $\mathbf{F}_{32 \times 32}$, we design a simple convolution target encoder $E_{t}$.
The $4 \times$ down-sampled $I_{tgt}^{\downarrow}$ is fed to $E_{t}$, then encoded features $\mathbf{F}_{32 \times 32}=E_{t}(I_{tgt}^{\downarrow})$ are conveyed to the StyleGAN $G$.

\noindent\textbf{Source Identity Encoder.}
Since $\textbf{w}_{8+}$ has the potential of injecting the source's identity information into the target without damaging the target's attributes, we map $I_{src}$ to the source identity embedding $\textbf{w}_{id}^{+} = E_{i}(I_{src}^{\downarrow})$ by using the source identity encoder $E_{i}$. 
Here, we utilize pSp encoder~\cite{psp} as the source identity encoder $E_{i}$ to map the overall source's identity attributes to $\mathcal{W+}$ space.

\noindent\textbf{Shape-Guided Identity Injection.}
Additionally, we exploit the 3DMM parameter space to focus on the source's structural information. 
To be specific, we leverage the 3DMM's shape parameter extracted from shape encoder $E_{s}$ which is a state-of-the-art 3DMM encoder~\cite{deca}.

Here, we only utilize the shape parameter, since the $G$ already has the capability of preserving the target image's poses by employing the $\textbf{F}_{32 \times 32}$.
Then, a mapping network $M: \mathcal{A} \rightarrow \mathcal{W}+$ produces $\mathbf{w}^{+}_{shape}$ with 3DMM's shape parameter $\alpha \in \mathcal{A}$.
\begin{equation}
    \mathbf{w}^{+}_{shape}=M(\alpha)=M(E_{s}(I_{src})).
\end{equation}
Finally, $\textbf{w}_{8+}$ is constructed with summation of shape embedding $\textbf{w}_{shape}^{+}$ and identity embedding $\textbf{w}_{id}^{+}$. 
Formally, 
\begin{equation}    
    \mathbf{w}_{8+}=\mathbf{w}_{shape}^{+}+\mathbf{w}_{id}^{+},
\end{equation}
where $\textbf{w}_{shape}^{+}$ is broadcast with the same size as $\textbf{w}_{id}^{+}$.

To summarize, our pipeline is described as 
\begin{equation}
    \hat{I}=G(\mathbf{F}_{32 \times 32},\textbf{w}_{8+}).
\label{eq:full}
\end{equation} 

\begin{table*}[t!]
    \centering
    \small
    \begin{adjustbox}{width=.75\linewidth,center}
    \begin{tabular}{l|c|c|c|c|c|c|c}
    \toprule
    \Xhline{2\arrayrulewidth}
    Methods &  Identity$\uparrow$ & Expression$\downarrow$ & Head Pose$\downarrow$ & Head Pose-HN$\downarrow$ & FID$\downarrow$ & Masked-L1$\downarrow$ & Eye Gazing$\downarrow$\\
    \Xhline{.8\arrayrulewidth}
    {SimSwap}   & 0.502 & 0.168 & \underline{0.016} &\underline{2.345} &34.84 & 0.046 &\underline{0.065} \\ 
    {InfoSwap}  & 0.557 & 0.196 & 0.021 & 3.533 & 15.75 & 0.067 & 0.068 \\ 
    {HifiFace}  & 0.515 & 0.210 & 0.021 & 3.486 & 30.91 & \underline{0.040} & 0.070 \\ 
    {MegaFS}    & 0.386 & 0.200 & 0.036 & 9.559 & 24.20 & 0.076 & 0.076 \\ 
    {FSLSD}     & 0.339 & 0.207 & 0.025 & 4.318 & \underline{12.41} & 0.046 & 0.081 \\ 
    {MFIM}      & \textbf{0.715} & \textbf{0.160} & 0.029 & 5.660 & 15.61 & 0.072 & 0.075 \\ 
    \hline
    Ours        & \underline{0.649} & \textbf{0.160} & \textbf{0.014} & \textbf{1.935} &\textbf{10.37} &\textbf{0.038} &\textbf{0.062} \\ 
    \Xhline{2\arrayrulewidth}
    \bottomrule
    \end{tabular}
    \end{adjustbox}
    \vspace{0.02cm}
    \caption{\textbf{Quantitative results} for comparison with baselines. \textbf{Bold} indicates the best score. \underline{Underline} indicates the second-best score. MFIM achieves the best identity score, but the reason for their handcrafted architecture could not prevent the \textbf{source attribute leakage} in respect of \textbf{Head Pose}, \textbf{Masked-L1} and \textbf{Eye Gazing}.
    \vspace{-0.4cm}
\label{table:baselines}}
\end{table*}

\subsection{Objective Functions}\label{three}

\noindent\textbf{Partial Landmark Loss.}
To encourage cooperation of the 3DMM's implicit and explicit information, we propose a partial landmark loss, only focusing on designated 51 landmarks out of 68 which supervise the source's inner facial shape. 
Moreover, such supervision also guides to more precise expression and head pose. 
To construct the ground truth of partial landmarks, we mix the target image's head pose and expression parameters and the source image's shape parameter and then feed the mixed parameters to the 3DMM decoder (\textit{i.e.}, FLAME~\cite{flame}) reconstructing the mesh, $Mesh^{mix}$, which is composed of 5023 vertices. 
More details are described in the supplementary materials. 

\begin{equation}
    \mathcal{L}_{pl} =\sum_{(i,j) \in Lmk} \norm{Mesh_{i}^{mix}-Mesh_{j}^{swap}},
    \label{formula:pl}
\end{equation}
where $Lmk$ is the set of inner face landmark pairs, $Mesh^{swap}$ represents swapped image's extracted mesh from $E_{s}$ and 3DMM decoder.

\noindent\textbf{Reconstruction Loss.}
We adopt the reconstruction loss for regularizing $\hat{I}$ attributes with $I_{tgt}$.
This part is composed of two losses, $L_{l2}$ and LPIPS~\cite{lpips} loss.
\begin{equation}
    \mathcal{L}_{recon}=\rVert{I_{tgt}-\hat{I}}_{1}\rVert_{2}+LPIPS(I_{tgt}, \hat{I})
\end{equation}

\noindent\textbf{Identity Loss.}
Identity loss ensures the $\hat{I}$ to have the same identity with $I_{src}$. 
\begin{equation}
    \mathcal{L}_{id}=1-\textrm{cossim}(R(I_{src}), R(\hat{I})),
\end{equation}
where $R$ is the pretrained face recognition model, ArcFace~\cite{arcface}. 
The notation cossim$(\cdot,\cdot)$ represents the cosine similarity between the ArcFace's embeddings.

\noindent\textbf{Adversarial Loss.}
Adversarial loss makes the model to generate the realistic $\hat{I}$.
We directly use the StyleGAN~\cite{sg2}'s non-saturating adversarial loss, $\mathcal{L}_{adv}$.
The detailed description is in supplementary materials.

\noindent\textbf{Total Objective.} 
\textbf{\ourmodel} is trained with the following total objective function: 
\begin{equation}
    \mathcal{L}_{total} = \lambda_{pl}\mathcal{L}_{pl} + \lambda_{recon}\mathcal{L}_{recon} + \lambda_{id}\mathcal{L}_{id} + \lambda_{adv}\mathcal{L}_{adv}, 
\end{equation}
where $\lambda_{pl}, \lambda_{recon}$, $\lambda_{id}$ and $\lambda_{adv}$ are the hyper-parameters. 

\section{Experiments}

\begin{figure*}[h!]
    \centering 
    \includegraphics[width=0.9\linewidth]{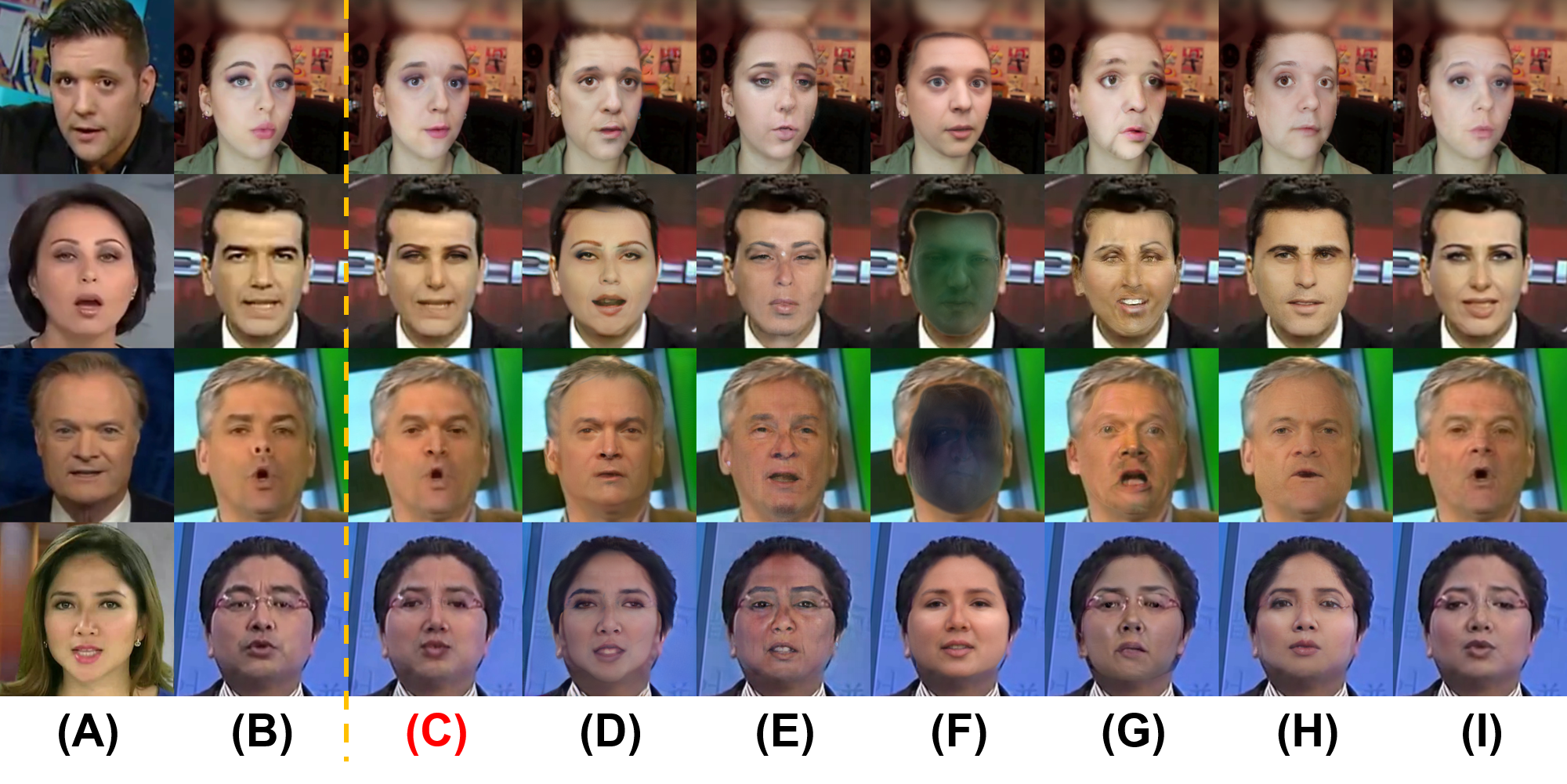}
    \vspace{-0.35cm}
    \caption{\textbf{Qualitative results} on $256 \times 256$ resolution FF++; (A) Source, (B) Target, \textbf{(C) \ourmodel}, (D) MFIM, (E) FSLSD, (F) MegaFS, (G) HifiFace, (H) InfoSwap, and (I) SimSwap. More results are in our supplementary materials. } 
    \vspace{-0.4cm}
\label{fig:FF++}
\end{figure*}

\begin{figure*}[t!]
    \centering 
    \includegraphics[width=0.9\textwidth]{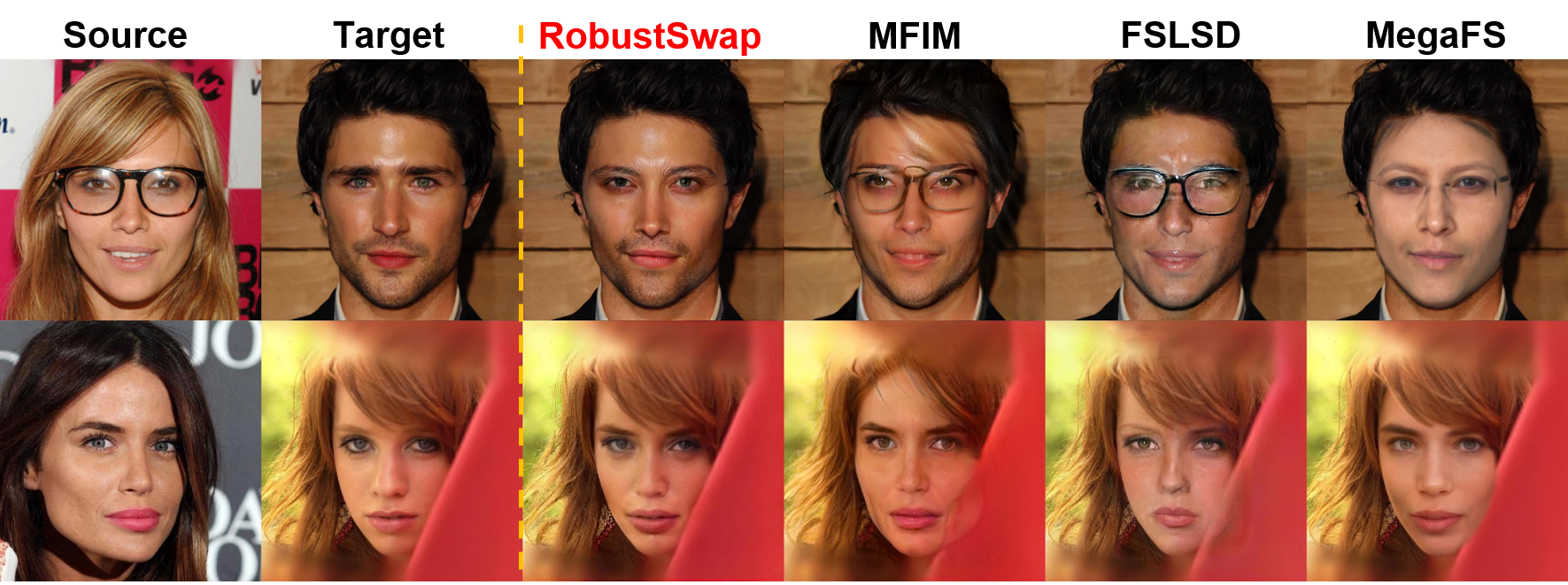} 
    \vspace{-0.2cm}
    \caption{\textbf{Qualitative results} on $1024 \times 1024$ resolution CelebA-HQ with megapixel baselines.} 
    \vspace{-0.6cm}
\label{fig:celeba}
\end{figure*}

\noindent\textbf{Datasets.} 
We train our model only on the FFHQ~\cite{sg1} without any identity-labeled or video datasets, different from previous methods~\cite{infoswap,styleswap,simswap,hififace}. 
We evaluate our method on CelebA-HQ~\cite{progan} validation set and FaceForensics++ (FF++)~\cite{ff++} dataset.
For CelebA-HQ, we sample 10,000 pairs randomly for the source and target images.
For FF++, we randomly select video pairs for qualitative evaluation. 
Note that the quantitative evaluations are only with CelebA-HQ, a high-resolution image dataset. 

\noindent\textbf{Baselines.} 
We compare our methods with the following face swapping baselines: SimSwap~\cite{simswap} InfoSwap~\cite{infoswap}, HifiFace~\cite{hififace}, MegaFS~\cite{megapixel}, FSLSD~\cite{fslsd} and MFIM~\cite{mfim}.
We utilize an unofficial code for HifiFace and reimplement the MFIM, strictly following the original paper.

\noindent\textbf{Implementation Details.} 
Our model is trained with 8 batch size on a NVIDIA A100 GPU for megapixels about 5 days. 
We use ADAM optimizer with a learning rate $1 \times 10^{-4}$. 
$\lambda_{id}$ and $\lambda_{rec}$ are set to 1.
$\lambda_{pl}$ is set to 100. $\lambda_{adv}$ is set to $10^{-2}$.

\subsection{Quantitative Evaluation}
\noindent\textbf{Evaluation Metrics.} 
For quantitative evaluation, the source and target image pairs are randomly sampled without duplication from CelebA-HQ~\cite{progan}.
We measure five metrics widely used for evaluating face swapping methods: Identity, Expression, Head Pose, Head Pose-HN, and Frechet Inception Distance (FID)~\cite{fid}.
In addition to five metrics, we employ two new metrics: \textbf{Masked-L1} and \textbf{Eye gazing}.
Identity score is the cosine similarity between the embedding vectors of $I_{src}$ and $\hat{I}$ extracted by a pre-trained face recognition model~\cite{cosface}, where we utilize a different model from the model used for the identity objective function.

Expression and Head Pose scores are calculated by measuring $L1$ distance between expression and head pose blendshape parameters of $I_{tgt}$ and $\hat{I}$ extracted by another pre-trained 3DMM encoder~\cite{ringnet}.
We measure Head Pose-HopeNet (HN) score by computing $L1$ distance between $I_{tgt}$ and $\hat{I}$ using a pre-trained head pose estimator~\cite{hopeneet}.
We also measure FID for the 10,000 $\hat{I}$ and real images of CelebA-HQ. 
\textbf{Masked-L1} measures the difference of the skin and head area excluding identity attributes between $\hat{I}$ and $I_{tgt}$. 
Specifically, we utilize a pre-trained face parsing map predictor~\cite{bisenet} for \textbf{Masked-L1} to extract only the skin and hair area of $I_{tgt}$ and $\hat{I}$.
Then, we measure the $L1$ distance for the pixels of the designated area. 
We employ \textbf{Masked-L1} to measure the \textbf{source attribute leakage} of appearances such as hair, glasses, and skin color.
Moreover, we utilize the pre-trained eye gazing estimator~\cite{rtgene} to evaluate the eye gazing of the swapped image $\hat{I}$.
In specific, we compute the $L1$ distance between eye gazing angles (\textit{e.g.}, yaw and pitch) of $I_{tgt}$ and $\hat{I}$.
The details for these two metrics are described in supplementary materials.

\begin{table}
\begin{center}
\begin{adjustbox}{width=1\columnwidth,center}







  \begin{tabular}{|l|l|l|l|l|l|l|}
    \hline
    \multirow{2}{*}{Method} &
      \multicolumn{2}{c}{ID sim \& Att pre$\uparrow$} &
      \multicolumn{2}{c}{Naturalness$\uparrow$} &
      \multicolumn{2}{c|}{Quality$\uparrow$} \\
    & Image & Video & Image & Video & Image & Video \\
    \hline
    HifiFace & 1.545           & 1.928          & 1.259             & 1.785 & 1.298 & 1.904 \\
    FSLSD & \underline{2.194} & 1.746           & \underline{2.311} & 1.642 & \underline{2.370} & 1.666 \\
    MFIM & 2.168               & \underline{2.500} & 1.857          & \underline{2.095} & 1.935 & \underline{2.190} \\
    \hline
    \textbf{\ourmodel} & \textbf{2.857} & \textbf{2.904} & \textbf{2.987} & \textbf{3.293} & \textbf{2.974} & \textbf{3.273} \\

    \hline
  \end{tabular}
\end{adjustbox}
\vspace{-0.4cm}
\end{center}
\caption{\textbf{User studies}. The larger score indicates the better, and the range of each criterion's score is set from 1 to 4.}
\label{table:user_study}
\vspace{-0.3cm}
\end{table}

\begin{table}
    \centering
    \begin{adjustbox}{width=1\columnwidth,center}
    \begin{tabular}{l|c|c|c}
    \Xhline{2.5\arrayrulewidth}
    Methods &  Identity$\uparrow$ & Expression$\downarrow$ & Head Pose$\downarrow$ \\
    \Xhline{1\arrayrulewidth}
    {Ours $8 \times 8$} & \textbf{0.684} & 0.223 & 0.026 \\
    {Ours $16 \times 16$} & 0.640 & 0.205 & 0.022 \\
    
    {Ours $32 \times 32$} & 0.620 & 0.184 & 0.018 \\
    {Ours $64 \times 64$} & 0.595 & \underline{0.166} & \underline{0.016} \\
    \hline
    {Ours full ($32 \times 32$)} & \underline{0.649} & \textbf{0.160} & \textbf{0.014} \\
    \Xhline{2.5\arrayrulewidth}
    \end{tabular}
    \end{adjustbox}
    \vspace{0.02cm}
    \caption{\textbf{Quantitative ablation studies}. Note that Ours full denotes the shape-guided identity injection and partial landmark loss added version.}
    \label{table:res_able}
    \vspace{-0.4cm}
\end{table}


\noindent\textbf{Comparison with Baselines.} 
Table~\ref{table:baselines} reports the quantitative comparison with baselines and \textbf{\ourmodel}.
\textbf{\ourmodel} model achieves the state-of-the-art performance compared to other face swapping baselines, except for the Identity score.
Although MFIM shows the best Identity score, the synthesized images of MFIM show severe \textbf{source attribute leakage} such as vanished hair and incorrect eye gazing and expression as shown in Fig.~\ref{fig:leakage}.

\noindent\textbf{Ablation Studies.}
As shown in Table~\ref{table:res_able}, we compare the performances of our model across different resolutions of $\mathbf{F}_{h \times w}$ (\textit{i.e.}, from $8 \times 8$ to $64 \times 64$).
Considering \textit{1st} to \textit{4th} row of our methods, there is the same tendency in Sec.~\ref{one} that the larger resolution of $\mathbf{F}_{h \times w}$, the lower expression, head pose errors, and identity score.
Since shape-guided identity injection and partial landmark loss boost the identity injection of the source image, Ours full achieves a higher identity score than Ours $16 \times 16$.
Lower expression and head pose scores demonstrate that the proposed techniques are also effective to preserve the target attributes.
 
\begin{figure*}[!t]
    \centering 
    \includegraphics[width=\textwidth]{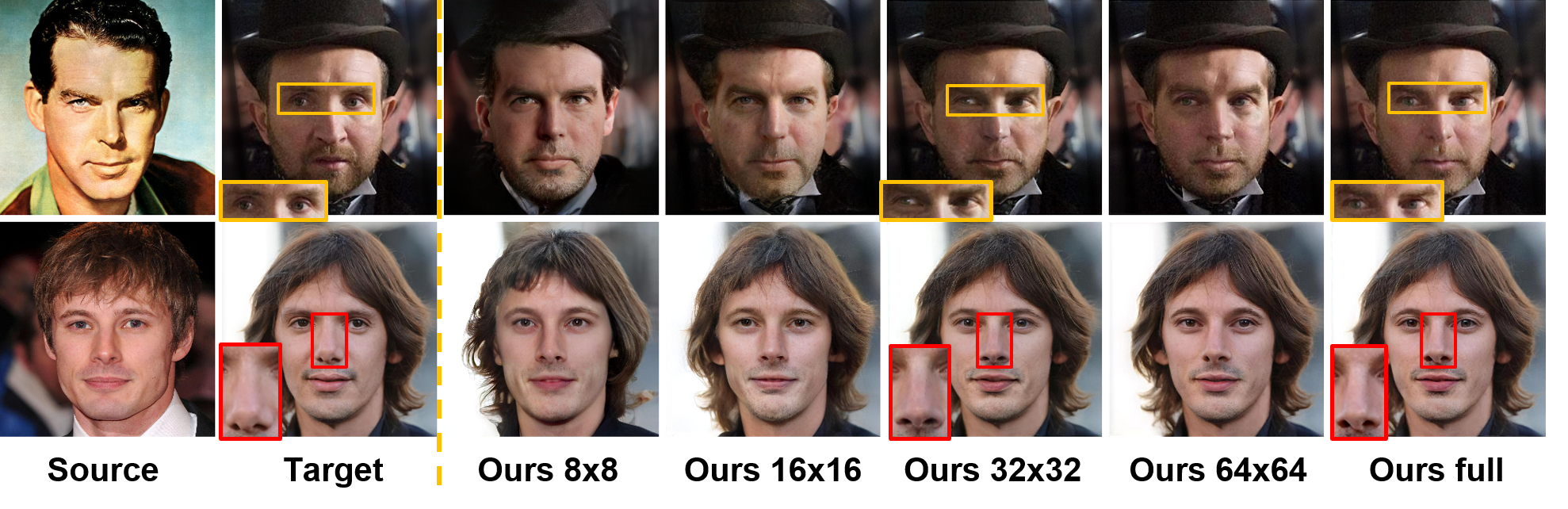}
    \vspace{-0.7cm}
    \caption{\textbf{Qualitative ablation studies} of $\mathbf{F}_{h \times w}$ and Ours full version. Ours $8 \times 8$ indicates that we embed target attributes to $\mathbf{F}_{8 \times 8}$ and use $\mathbf{w}_{4+}$ for injecting identity attributes of a source. Please be aware of the \textcolor{Dandelion}{yellow} and \textcolor{red}{red} boxes. Ours full is improved in respects of preserving the target image's pose such as eye gazing and expression, and reflecting the source image's shape compared with Ours $32 \times 32$.}
    \vspace{-0.3cm}
\label{fig:res_ablation}
\end{figure*}

\begin{figure}[t!]
    \centering 
    \includegraphics[width=\linewidth]{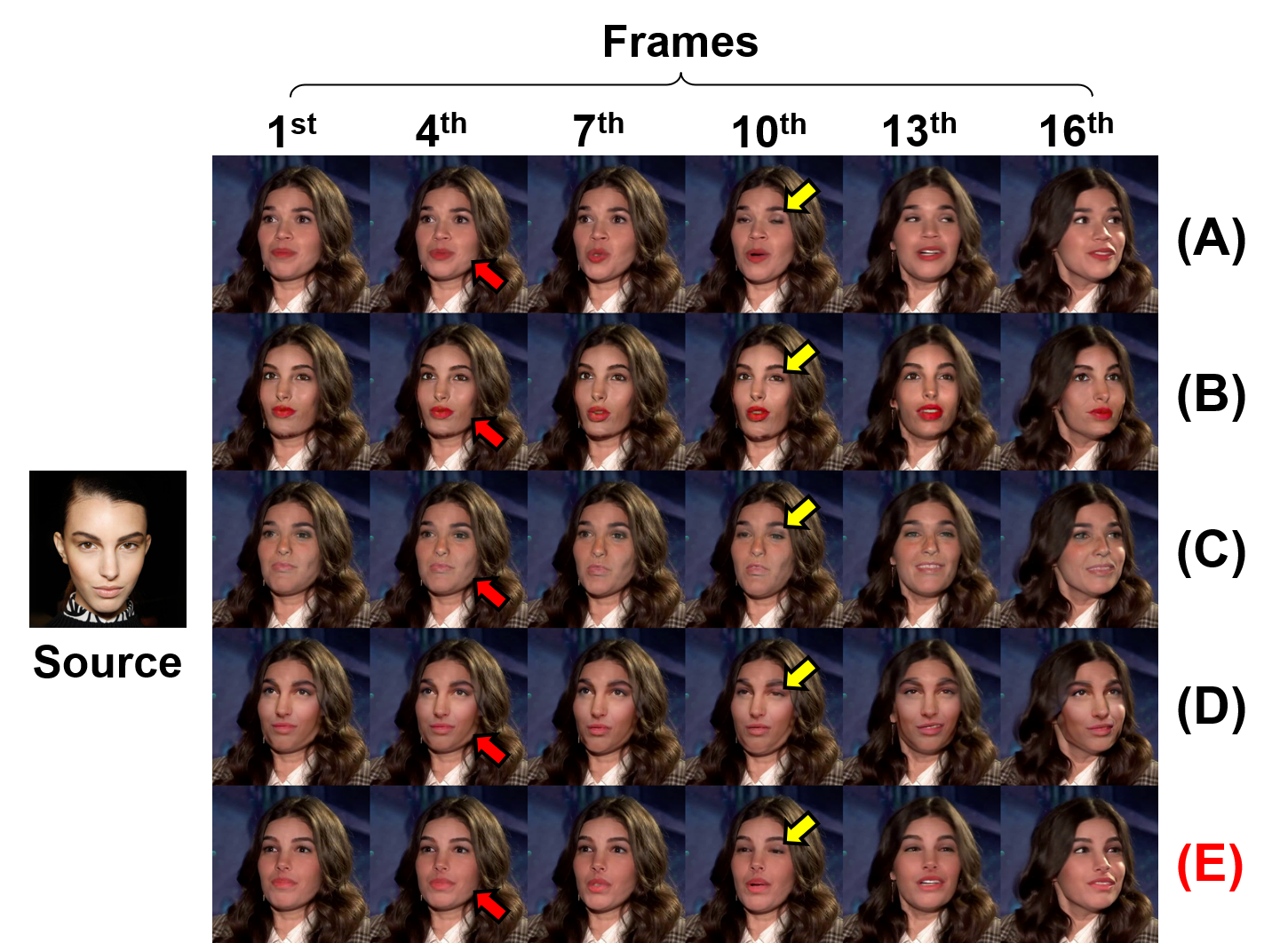} 
    \vspace{-0.5cm}
    \caption{\textbf{Qualitative results} on $512 \times 512$ resolution CelebV-HQ; (A) Target frames, (B) HifiFace, (C) FSLSD, (D) MFIM, \textbf{(E) RobustSwap}. Please be aware of the \textcolor{Dandelion}{yellow} and \textcolor{red}{red} arrows. Baselines suffer from \textbf{source attribute leakage}.}
    \vspace{-0.5cm}
\label{fig:vid}
\end{figure}

\noindent\textbf{User Studies.} 
We further evaluate our model and three recent baselines~\cite{hififace,fslsd,mfim} via a user study on synthesizing the images and the videos. 
The participants evaluated 11 swapped image samples from CelebA-HQ and 6 video samples from CelebV-HQ~\cite{celebvhq}.
The users are asked to score the quality of swapped images and videos according to the following criteria: 1) Identity similarity and Attribute preservation (ID sim \& Att pre); 2) Naturalness; and 3) Quality. 
We designate the highest score to be 4 and the lowest score to be 1 for each criterion. Table~\ref{table:user_study} shows that our method achieves the best score in every criterion, demonstrating that our results are the most plausible in human perceptual evaluation.
Notably, our video score is higher than other baselines with large margin even though we do not train any video datasets. 
These results indicate that preventing the \textbf{source attribute leakage} is also crucial for synthesizing temporally consistent videos.

\subsection{Qualitative Evaluation}

\noindent\textbf{Comparison of Baselines.}
We compare our \textbf{\ourmodel} and baselines on CelebA-HQ and FF++ datasets.
As shown in Fig.~\ref{fig:FF++}, SimSwap sometimes shows \textbf{source attribute leakage} such as bringing the source's hair lines with low-quality results.
InfoSwap and HifiFace often fail to retain the target image's expression and eye gazing.
In contrast, our method generates more perceptually convincing swapped images without source \textbf{source attribute leakage}.

In Fig.~\ref{fig:celeba}, we compare our \textbf{\ourmodel} with megapixel models~\cite{mfim,fslsd,megapixel}. 
Although MFIM and FSLSD reflect source identity well, they often produce visual artifacts like \textit{appearance leakage} (\textit{e.g.}, hairstyle and eyeglasses) and \textit{pose leakage} (\textit{e.g.}, incorrect eye gazing and expression).
MegaFS generates inaccurate skin-colored images. 
On the other hand, our method robustly changes the target face to the source's one almost without \textbf{source attribute leakage}, following the target image's eye gazing and expression, and having no texture leakage from the source image.
Moreover, as shown in Fig.~\ref{fig:vid}, we compare the recent three baselines with our method on generating the videos. 
While the baselines show the \textit{pose} and \textit{appearance leakages}, \textbf{\ourmodel} is robust to \textit{source attribute leakage} even in the video.
Notably, these results show that \textit{source attribute leakage} is also crucial to synthesize the temporally consistent videos in face swapping task.

\noindent\textbf{Ablation Studies.}
As shown in Fig.~\ref{fig:res_ablation}, Ours $8 \times 8$ and Ours $16 \times 16$ hardly preserve the hat, hairstyle, skin color, eye gazing, and expression of the target image.  
In contrast, Ours $64 \times 64$ faithfully follows the appearance and pose of the target, but the identity of the result is quite heterogeneous with the source. 
Ours $32 \times 32$ contains the target's attribute and source's identity in balance, which demonstrates that our analysis in Sec.~\ref{one} is effective for searching the proper combination of latent spaces.
While the performance of Ours $32 \times 32$ is commendable, there is a room for improvement in accurately preserving the pose of the inner face region.
Therefore, by leveraging the shape-guided identity injection and partial landmark loss, Ours full can preserve more detailed expression and head pose, and simultaneously reflect source's inner shape than Ours $32 \times 32$. 


\section{Conclusion}
In this paper, we propose a robust face swapping model, \textbf{\ourmodel}, which solves \textbf{source attribute leakage} problems. 
We analyze the latent space of StyleGAN for face swapping, ultimately we develop a simple yet robust face swapping model without any architectural modification of StyleGAN, which is easy to train and implement.
On the other hand, we believe that our model can be extended to other combinations of subspaces, not limited to only face swapping tasks. 
We further utilize the explicit and implicit information of 3DMM to provide more detailed source identity information and precise target person's pose. 
Our experiments show that \textbf{\ourmodel} is comparable with previous face swapping models. Additionally, \textbf{\ourmodel} shows high-quality results in video face swapping without video datasets. 
We believe that our analysis on StyleGAN for face swapping inspires the future researchers to analyze the latent spaces of the generative model in perspective of face swapping task and utilize it as a strong prior for face swapping.


{\small
\bibliographystyle{ieee_fullname}
\bibliography{reference}

\begin{thebibliography}{10}\itemsep=-1pt

\bibitem{image2stylegan}
Rameen Abdal, Yipeng Qin, and Peter Wonka.
\newblock Image2stylegan: How to embed images into the stylegan latent space?
\newblock In {\em Proceedings of the IEEE/CVF International Conference on
  Computer Vision}, pages 4432--4441, 2019.

\bibitem{image2stylegan++}
Rameen Abdal, Yipeng Qin, and Peter Wonka.
\newblock Image2stylegan++: How to edit the embedded images?
\newblock In {\em Proceedings of the IEEE/CVF conference on computer vision and
  pattern recognition}, pages 8296--8305, 2020.

\bibitem{ls3dmm}
James Booth, Anastasios Roussos, Allan Ponniah, David Dunaway, and Stefanos
  Zafeiriou.
\newblock Large scale 3d morphable models.
\newblock {\em International Journal of Computer Vision}, 126(2):233--254,
  2018.

\bibitem{vggface2}
Qiong Cao, Li Shen, Weidi Xie, Omkar~M Parkhi, and Andrew Zisserman.
\newblock Vggface2: A dataset for recognising faces across pose and age.
\newblock In {\em 2018 13th IEEE international conference on automatic face \&
  gesture recognition (FG 2018)}, pages 67--74. IEEE, 2018.

\bibitem{simswap}
Renwang Chen, Xuanhong Chen, Bingbing Ni, and Yanhao Ge.
\newblock Simswap: An efficient framework for high fidelity face swapping.
\newblock In {\em Proceedings of the 28th ACM International Conference on
  Multimedia}, pages 2003--2011, 2020.

\bibitem{voxceleb2}
Joon~Son Chung, Arsha Nagrani, and Andrew Zisserman.
\newblock Voxceleb2: Deep speaker recognition.
\newblock {\em arXiv preprint arXiv:1806.05622}, 2018.

\bibitem{arcface}
Jiankang Deng, Jia Guo, Niannan Xue, and Stefanos Zafeiriou.
\newblock Arcface: Additive angular margin loss for deep face recognition.
\newblock In {\em Proceedings of the IEEE/CVF conference on computer vision and
  pattern recognition}, pages 4690--4699, 2019.

\bibitem{deep3drecon}
Yu Deng, Jiaolong Yang, Sicheng Xu, Dong Chen, Yunde Jia, and Xin Tong.
\newblock Accurate 3d face reconstruction with weakly-supervised learning: From
  single image to image set.
\newblock In {\em Proceedings of the IEEE/CVF conference on computer vision and
  pattern recognition workshops}, pages 0--0, 2019.

\bibitem{deca}
Yao Feng, Haiwen Feng, Michael~J Black, and Timo Bolkart.
\newblock Learning an animatable detailed 3d face model from in-the-wild
  images.
\newblock {\em ACM Transactions on Graphics (ToG)}, 40(4):1--13, 2021.

\bibitem{rtgene}
Tobias Fischer, Hyung~Jin Chang, and Yiannis Demiris.
\newblock Rt-gene: Real-time eye gaze estimation in natural environments.
\newblock In {\em Proceedings of the European conference on computer vision
  (ECCV)}, pages 334--352, 2018.

\bibitem{infoswap}
Gege Gao, Huaibo Huang, Chaoyou Fu, Zhaoyang Li, and Ran He.
\newblock Information bottleneck disentanglement for identity swapping.
\newblock In {\em Proceedings of the IEEE/CVF conference on computer vision and
  pattern recognition}, pages 3404--3413, 2021.

\bibitem{fid}
Martin Heusel, Hubert Ramsauer, Thomas Unterthiner, Bernhard Nessler, and Sepp
  Hochreiter.
\newblock Gans trained by a two time-scale update rule converge to a local nash
  equilibrium.
\newblock {\em Advances in neural information processing systems}, 30, 2017.

\bibitem{bfm}
IEEE.
\newblock {\em A 3D Face Model for Pose and Illumination Invariant Face
  Recognition}, Genova, Italy, 2009.

\bibitem{oorinversion}
Kyoungkook Kang, Seongtae Kim, and Sunghyun Cho.
\newblock Gan inversion for out-of-range images with geometric transformations.
\newblock In {\em Proceedings of the IEEE/CVF International Conference on
  Computer Vision}, pages 13941--13949, 2021.

\bibitem{progan}
Tero Karras, Timo Aila, Samuli Laine, and Jaakko Lehtinen.
\newblock Progressive growing of gans for improved quality, stability, and
  variation.
\newblock {\em arXiv preprint arXiv:1710.10196}, 2017.

\bibitem{sg3}
Tero Karras, Miika Aittala, Samuli Laine, Erik H{\"a}rk{\"o}nen, Janne
  Hellsten, Jaakko Lehtinen, and Timo Aila.
\newblock Alias-free generative adversarial networks.
\newblock {\em Advances in Neural Information Processing Systems}, 34:852--863,
  2021.

\bibitem{sg1}
Tero Karras, Samuli Laine, and Timo Aila.
\newblock A style-based generator architecture for generative adversarial
  networks.
\newblock In {\em Proceedings of the IEEE/CVF conference on computer vision and
  pattern recognition}, pages 4401--4410, 2019.

\bibitem{sg2}
Tero Karras, Samuli Laine, Miika Aittala, Janne Hellsten, Jaakko Lehtinen, and
  Timo Aila.
\newblock Analyzing and improving the image quality of stylegan.
\newblock In {\em Proceedings of the IEEE/CVF conference on computer vision and
  pattern recognition}, pages 8110--8119, 2020.

\bibitem{styleyourhair}
Taewoo Kim, Chaeyeon Chung, Yoonseo Kim, Sunghyun Park, Kangyeol Kim, and
  Jaegul Choo.
\newblock Style your hair: Latent optimization for pose-invariant hairstyle
  transfer via local-style-aware hair alignment.
\newblock In {\em Computer Vision--ECCV 2022: 17th European Conference, Tel
  Aviv, Israel, October 23--27, 2022, Proceedings, Part XVII}, pages 188--203.
  Springer, 2022.

\bibitem{faceshifter}
Lingzhi Li, Jianmin Bao, Hao Yang, Dong Chen, and Fang Wen.
\newblock Faceshifter: Towards high fidelity and occlusion aware face swapping.
\newblock {\em arXiv preprint arXiv:1912.13457}, 2019.

\bibitem{flame}
Tianye Li, Timo Bolkart, Michael.~J. Black, Hao Li, and Javier Romero.
\newblock Learning a model of facial shape and expression from {4D} scans.
\newblock {\em ACM Transactions on Graphics, (Proc. SIGGRAPH Asia)},
  36(6):194:1--194:17, 2017.

\bibitem{mfim}
Sanghyeon Na.
\newblock Mfim: Megapixel facial identity manipulation.
\newblock In {\em Computer Vision--ECCV 2022: 17th European Conference, Tel
  Aviv, Israel, October 23--27, 2022, Proceedings, Part XIII}, pages 143--159.
  Springer, 2022.

\bibitem{voxceleb1}
Arsha Nagrani, Joon~Son Chung, and Andrew Zisserman.
\newblock Voxceleb: a large-scale speaker identification dataset.
\newblock {\em arXiv preprint arXiv:1706.08612}, 2017.

\bibitem{vggface1}
Omkar~M. Parkhi, Andrea Vedaldi, and Andrew Zisserman.
\newblock Deep face recognition.
\newblock In {\em Proceedings of the British Machine Vision Conference (BMVC)},
  pages 41.1--41.12. BMVA Press, September 2015.

\bibitem{psp}
Elad Richardson, Yuval Alaluf, Or Patashnik, Yotam Nitzan, Yaniv Azar, Stav
  Shapiro, and Daniel Cohen-Or.
\newblock Encoding in style: a stylegan encoder for image-to-image translation.
\newblock In {\em Proceedings of the IEEE/CVF conference on computer vision and
  pattern recognition}, pages 2287--2296, 2021.

\bibitem{ff++}
Andreas Rossler, Davide Cozzolino, Luisa Verdoliva, Christian Riess, Justus
  Thies, and Matthias Nie{\ss}ner.
\newblock Faceforensics++: Learning to detect manipulated facial images.
\newblock In {\em Proceedings of the IEEE/CVF international conference on
  computer vision}, pages 1--11, 2019.

\bibitem{hopeneet}
Nataniel Ruiz, Eunji Chong, and James~M Rehg.
\newblock Fine-grained head pose estimation without keypoints.
\newblock In {\em Proceedings of the IEEE conference on computer vision and
  pattern recognition workshops}, pages 2074--2083, 2018.

\bibitem{ringnet}
Soubhik Sanyal, Timo Bolkart, Haiwen Feng, and Michael~J Black.
\newblock Learning to regress 3d face shape and expression from an image
  without 3d supervision.
\newblock In {\em Proceedings of the IEEE/CVF Conference on Computer Vision and
  Pattern Recognition}, pages 7763--7772, 2019.

\bibitem{e4e}
Omer Tov, Yuval Alaluf, Yotam Nitzan, Or Patashnik, and Daniel Cohen-Or.
\newblock Designing an encoder for stylegan image manipulation.
\newblock {\em ACM Transactions on Graphics (TOG)}, 40(4):1--14, 2021.

\bibitem{cosface}
Hao Wang, Yitong Wang, Zheng Zhou, Xing Ji, Dihong Gong, Jingchao Zhou, Zhifeng
  Li, and Wei Liu.
\newblock Cosface: Large margin cosine loss for deep face recognition.
\newblock In {\em Proceedings of the IEEE conference on computer vision and
  pattern recognition}, pages 5265--5274, 2018.

\bibitem{wang2022high}
Tengfei Wang, Yong Zhang, Yanbo Fan, Jue Wang, and Qifeng Chen.
\newblock High-fidelity gan inversion for image attribute editing.
\newblock In {\em Proceedings of the IEEE/CVF Conference on Computer Vision and
  Pattern Recognition}, pages 11379--11388, 2022.

\bibitem{hififace}
Yuhan Wang, Xu Chen, Junwei Zhu, Wenqing Chu, Ying Tai, Chengjie Wang, Jilin
  Li, Yongjian Wu, Feiyue Huang, and Rongrong Ji.
\newblock Hififace: 3d shape and semantic prior guided high fidelity face
  swapping.
\newblock {\em arXiv preprint arXiv:2106.09965}, 2021.

\bibitem{uniswap}
Chao Xu, Jiangning Zhang, Yue Han, Guanzhong Tian, Xianfang Zeng, Ying Tai,
  Yabiao Wang, Chengjie Wang, and Yong Liu.
\newblock Designing one unified framework for high-fidelity face reenactment
  and swapping.
\newblock In {\em Computer Vision--ECCV 2022: 17th European Conference, Tel
  Aviv, Israel, October 23--27, 2022, Proceedings, Part XV}, pages 54--71.
  Springer, 2022.

\bibitem{fslsd}
Yangyang Xu, Bailin Deng, Junle Wang, Yanqing Jing, Jia Pan, and Shengfeng He.
\newblock High-resolution face swapping via latent semantics disentanglement.
\newblock In {\em Proceedings of the IEEE/CVF Conference on Computer Vision and
  Pattern Recognition}, pages 7642--7651, 2022.

\bibitem{styleswap}
Zhiliang Xu, Hang Zhou, Zhibin Hong, Ziwei Liu, Jiaming Liu, Zhizhi Guo, Junyu
  Han, Jingtuo Liu, Errui Ding, and Jingdong Wang.
\newblock Styleswap: Style-based generator empowers robust face swapping.
\newblock In {\em Computer Vision--ECCV 2022: 17th European Conference, Tel
  Aviv, Israel, October 23--27, 2022, Proceedings, Part XIV}, pages 661--677.
  Springer, 2022.

\bibitem{styleheat}
Fei Yin, Yong Zhang, Xiaodong Cun, Mingdeng Cao, Yanbo Fan, Xuan Wang, Qingyan
  Bai, Baoyuan Wu, Jue Wang, and Yujiu Yang.
\newblock Styleheat: One-shot high-resolution editable talking face generation
  via pre-trained stylegan.
\newblock In {\em Computer Vision--ECCV 2022: 17th European Conference, Tel
  Aviv, Israel, October 23--27, 2022, Proceedings, Part XVII}, pages 85--101.
  Springer, 2022.

\bibitem{bisenet}
Changqian Yu, Jingbo Wang, Chao Peng, Changxin Gao, Gang Yu, and Nong Sang.
\newblock Bisenet: Bilateral segmentation network for real-time semantic
  segmentation.
\newblock In {\em Proceedings of the European conference on computer vision
  (ECCV)}, pages 325--341, 2018.

\bibitem{lpips}
Richard Zhang, Phillip Isola, Alexei~A Efros, Eli Shechtman, and Oliver Wang.
\newblock The unreasonable effectiveness of deep features as a perceptual
  metric.
\newblock In {\em Proceedings of the IEEE conference on computer vision and
  pattern recognition}, pages 586--595, 2018.

\bibitem{celebvhq}
Hao Zhu, Wayne Wu, Wentao Zhu, Liming Jiang, Siwei Tang, Li Zhang, Ziwei Liu,
  and Chen~Change Loy.
\newblock {CelebV-HQ}: A large-scale video facial attributes dataset.
\newblock In {\em ECCV}, 2022.

\bibitem{barbershop}
Peihao Zhu, Rameen Abdal, John Femiani, and Peter Wonka.
\newblock Barbershop: Gan-based image compositing using segmentation masks.
\newblock {\em arXiv preprint arXiv:2106.01505}, 2021.

\bibitem{megapixel}
Yuhao Zhu, Qi Li, Jian Wang, Cheng-Zhong Xu, and Zhenan Sun.
\newblock One shot face swapping on megapixels.
\newblock In {\em Proceedings of the IEEE/CVF conference on computer vision and
  pattern recognition}, pages 4834--4844, 2021.

\end{thebibliography}
}

\clearpage

\twocolumn[
\begin{center}
    \vspace*{1.1cm}
    \Large{\bf{Supplementary Material}}
    \vspace*{1.7cm}
\end{center}]

\section*{A. Architecture}
This section outlines the details of architectures for the target attribute encoder, source identity encoder, source shape encoder, mapper, and generator which are described in Table ~\ref{supp_table:architecture}.

\begin{table*}
\begin{center}
\begin{adjustbox}{width=\linewidth}

  \begin{tabular}{|l|l|l|l|}
    \hline
  
    Module & Input $\rightarrow$ Output & Layer  \\\Xhline{2.5\arrayrulewidth}
    \multirow{3}{*}{Target Attributes Encoder $E_{t}$} & \multirow{2}{*}{$I_{tgt}^{\downarrow}$ $\rightarrow$ $\mathbf{F}_{32 \times 32}$} & Conv(3,1,1) $\rightarrow$ LeakyReLU $\rightarrow$ Conv(3,1,1) $\rightarrow$ LeakyReLU $\rightarrow$ \\
                                     &                                                                  &$\{$ Conv(3,2,1) $\rightarrow$ LeakyReLU $\rightarrow$ Conv(3,1,1) $\rightarrow$ LeakyReLU $\}$ $\times$4 \\
                                     & $\mathbf{F}_{32 \times 32} \rightarrow \mathbf{I}_{32 \times 32}$& Conv(1,1,1) \\\hline
    Source Identity Encoder $E_{i}$  & $I_{src}^{\downarrow}$ $\rightarrow$ $\textbf{w}_{id}^{+}$       & pSp~\cite{psp} Encoder    \\\hline
    Source Shape Encoder $E_{s}$     & $I_{src}^{\downarrow}$ $\rightarrow$ $\alpha$                    & DECA~\cite{deca} Encoder    \\\hline  
    Mapper $M$                       & $\alpha$ $\rightarrow$ $\mathbf{w}_{shape}^{+}$                  & 5 EqualLinear Layers with LeakyReLU \\\hline  
    \multirow{2}{*}{$h \times w$ Block}               & $\mathbf{F}_{h/2 \times w/2}$, $\{\textbf{w}_{i}, \textbf{w}_{i+1}\}$ $\rightarrow$ $\mathbf{F}_{h \times w}$        & Upsample $\rightarrow$ StyleConv $\rightarrow$ NoiseInjection $\rightarrow$ StyleConv $\rightarrow$ NoiseInjection \\
                                     & $\mathbf{F}_{h \times w}$, $\mathbf{I}_{h/2 \times w/2}$, $\textbf{w}_{i+2}$ $\rightarrow$ $\mathbf{I}_{h \times w}$ & ToRGB \\ \hline
    
    \multirow{5}{*}{Generator $G$} 
    & $\mathbf{F}_{32 \times 32}$, $\mathbf{I}_{32 \times 32}$,     $\{\textbf{w}_{8}, \textbf{w}_{9}, \textbf{w}_{10}\}$ $\rightarrow$     $\mathbf{F}_{64 \times 64}$, $\mathbf{I}_{64 \times 64}$     & 64  $\times$ 64 Block \\
    & $\mathbf{F}_{64 \times 64}$, $\mathbf{I}_{64 \times 64}$,     $\{\textbf{w}_{10}, \textbf{w}_{11}, \textbf{w}_{12}\}$ $\rightarrow$   $\mathbf{F}_{128 \times 128}$, $\mathbf{I}_{128 \times 128}$ & 128 $\times$ 128 Block \\
    & $\mathbf{F}_{128 \times 128}$, $\mathbf{I}_{128 \times 128}$, $\{\textbf{w}_{12}, \textbf{w}_{13}, \textbf{w}_{14}\}$ $\rightarrow$   $\mathbf{F}_{256 \times 256}$, $\mathbf{I}_{256 \times 256}$ & 256 $\times$ 256 Block \\
    & $\mathbf{F}_{256 \times 256}$, $\mathbf{I}_{256 \times 256}$, $\{\textbf{w}_{14}, \textbf{w}_{15}, \textbf{w}_{16}\}$ $\rightarrow$   $\mathbf{F}_{512 \times 512}$, $\mathbf{I}_{512 \times 512}$ & 512 $\times$ 512 Block \\
    & $\mathbf{F}_{512 \times 512}$, $\mathbf{I}_{512 \times 512}$, $\{\textbf{w}_{16}, \textbf{w}_{17}, \textbf{w}_{18}\}$ $\rightarrow$   $\mathbf{F}_{1024 \times 1024}$, $\hat{I}$                   & 1024 $\times$ 1024 Block \\

    \hline
  \end{tabular}
\end{adjustbox}
\end{center}
\caption{Architecture details of \textbf{\ourmodel}. The target attributes encoder $E_{t}$ maps the spatial information of the down-sampled target image $I_{tgt}^{\downarrow} \in \mathbb{R}^{256 \times 256 \times 3}$ to the feature map $\mathbf{F}_{32 \times 32} \in \mathbb{R}^{32 \times 32 \times 512}$ and the low resolution image $\mathbf{I}_{32 \times 32}  \in \mathbb{R}^{32 \times 32 \times 3}$. Here, Conv(k,s,p) denotes a 2D Convolutional layer with kernel size k, stride size s, and padding size p. The source identity encoder $E_{i}$ and, the source shape encoder and mapper ($E_{s}$, $M$) map the identity information of the down-sampled source image $I_{src}^{\downarrow} \in \mathbb{R}^{256 \times 256 \times 3}$ to $\mathbf{w}$ vectors, $\mathbf{w}_{id}^{+}$ and $\mathbf{w}_{shape}^{+}$, respectively. In $h \times w$ Block, StyleConv, NoiseInjection, and ToRGB are exactly the same as in StyleGAN~\cite{sg2}. The generator $G$ consists of multiple $h \times w$ Blocks to generate the swapped image $\hat{I} \in \mathbb{R}^{1024 \times 1024 \times 3}$. } 
\label{supp_table:architecture}

\end{table*}

 
\section*{B. Details of StyleGAN's $\mathcal{F/W+}$ Analysis}
As already described in Sec.~3 of our main paper, we conduct the depth analysis to investigate the conformity of StyleGAN~\cite{sg2}'s latent combinations for face swapping task. 
Following subsections describe the more detailed analysis.

\begin{figure}[h!]
    \centering 
    \includegraphics[width=\linewidth]{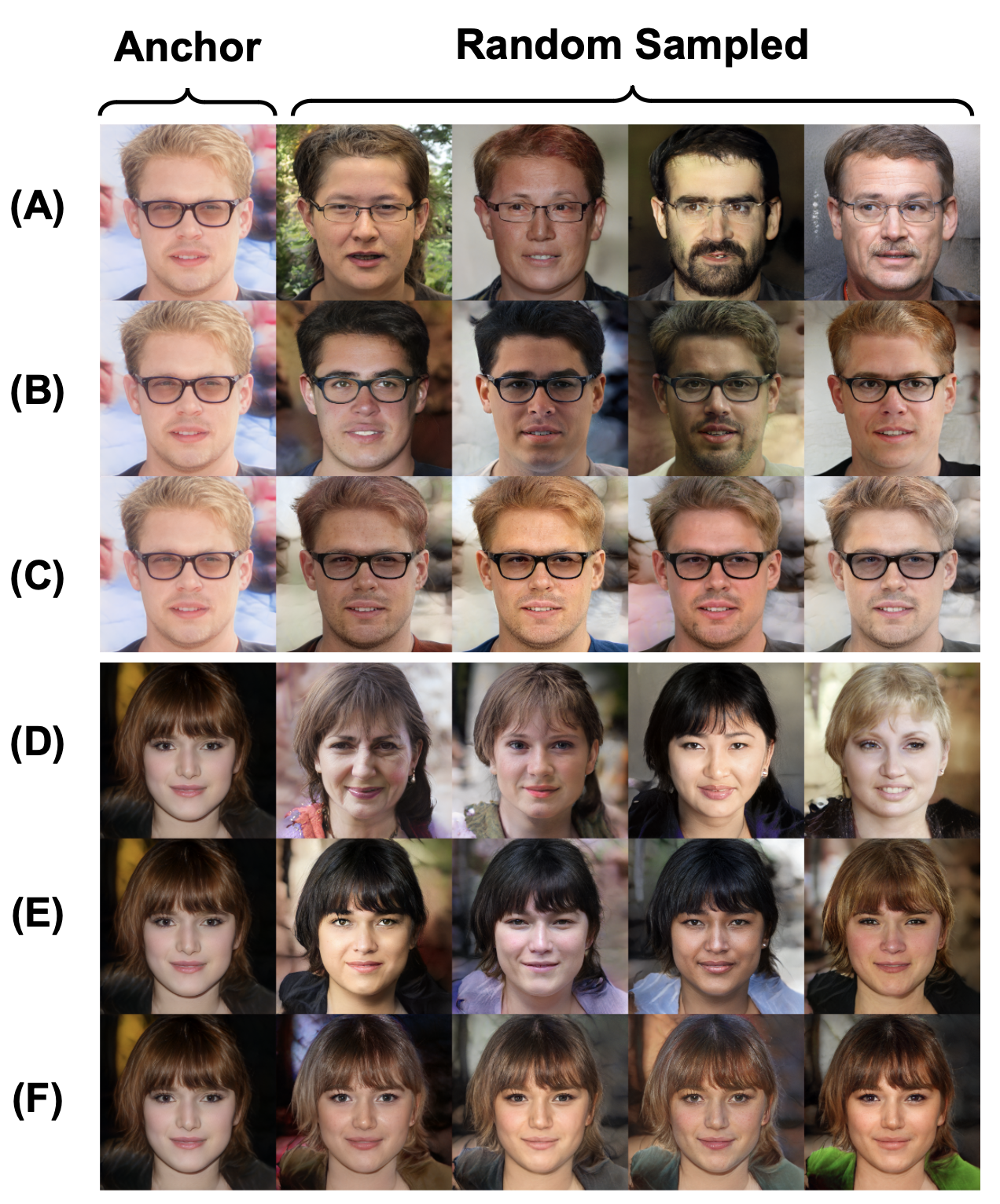}
    \caption{\textbf{Qualitative analysis.} More examples from the StyleGAN latent analysis for face swapping task. 
    An \textit{anchor} image is obtained from the inverted vectors $\textbf{w}_{1+}$ by using GAN inversion method~\cite{psp}. \textit{Random sampled} images of (A) and (D) are generated by the fixed feature map $\mathbf{F}^{*}_{16 \times 16}$ and randomly initialized $\mathbf{w}_{6+}$. 
    (B) and (E)'s \textit{random sampled} images are produced by $\mathbf{F}^{*}_{32 \times 32}$ and $\mathbf{w}_{8+}$.
    \textit{Random sampled} images of (C) and (F) are obtained from $\mathbf{F}^{*}_{64 \times 64}$ and $\mathbf{w}_{10+}$.} 
    \label{fig:quali}
\end{figure}

\subsection*{B.1. Metrics for Quantitative Analysis.}
To find the appropriate combination of latents to maintain the identity-irrelevant attributes while modifying identity information, we quantitatively compare the \textit{anchor image} (correspond to a target image in face swapping) and other \textit{random sampled} images (correspond to swapped images in face swapping) with four metrics. 
Please see Table \ref{supp_table:analysis_spec}, which contains the specification of our quantitative analysis factors.

\subsection*{B.2. More Qualitative Results of the analysis.}
Furthermore, we conduct the qualitative analysis on the three highest overall scored combinations ($\mathbf{F}_{16 \times 16}$, $\mathbf{w}_{6}^{+}$), ($\mathbf{F}_{32 \times 32}$, $\mathbf{w}_{8}^{+}$) and ($\mathbf{F}_{64 \times 64}$, $\mathbf{w}_{10}^{+}$).
In this section, we show more qualitative results which are not shown in the main paper due to limited space. 
In Fig.~\ref{fig:quali} (B) and (E), compared with the \textit{anchor} image, other \textit{random sampled} images' eyeglasses and hair bang are preserved while changing identity. 
On the other hand, in (A) and (D), those appearance attributes such as eyeglasses and hair vary. 
Moreover, in the (C) and (F), except for the light condition and skin color, there are almost no changes. 
Therefore, we choose the combination of ($\mathbf{F}_{32 \times 32}$, $\mathbf{w}_{8}^{+}$) since it properly preserves pose and appearance attributes and can change the capability of identity.


\section*{C. New Metrics}
Following two new metrics are proposed in this work for measuring the more precise degree of \textit{source attribute leakage}.

\subsection*{C.1. Eye gazing.}
\textbf{Eye gazing} metric is calculated by RT-GENE~\cite{rtgene}'s yaw and pitch $L1$ error between the target and swapped images.
This metric helps to estimate the part of \textit{pose leakage}. 

\subsection*{C.2. Masked-L1.}
\begin{figure}[t!]
    \centering 
    \includegraphics[width=\linewidth]{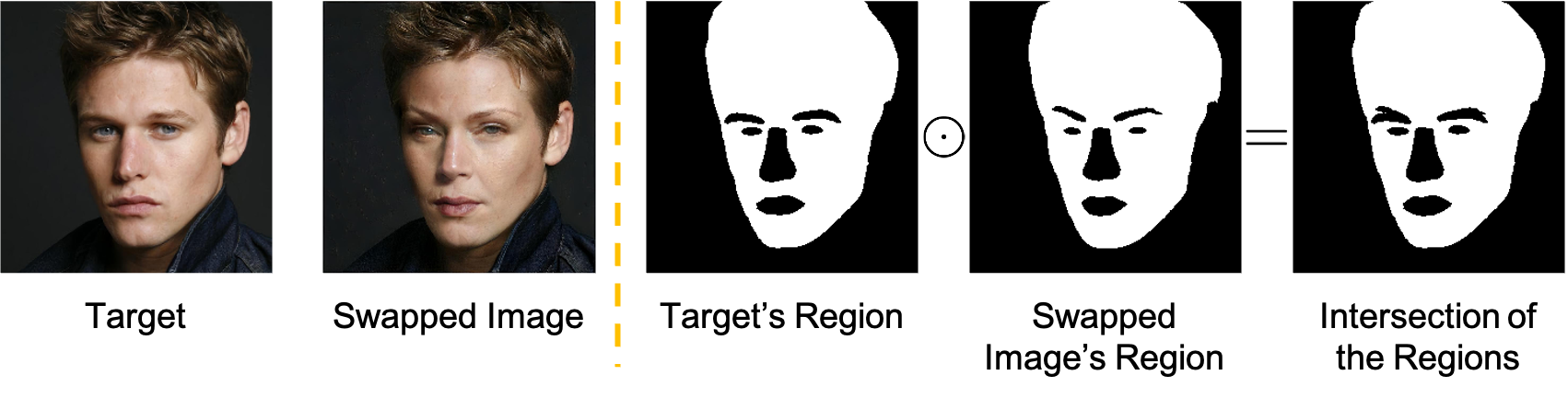}
    \caption{\textbf{Logic of the measurement of Masked-L1 metric.} Since the target and swapped images have different eyes, nose, lip and eyebrows, we exclude those region for calculating the metric score. Moreover, for measuring on the shared region, we designate the region as intersection.}
    \label{fig:mask1l}
\end{figure}

\textbf{Masked-L1} measures the error of the hair and skin region between the target and swapped images.
We utilize the off-the-shelf face parsing map predictor BiseNet~\cite{bisenet} for extracting the region.
Note that as can be seen in Fig.~\ref{fig:mask1l}, for excluding the other region, the error is calculated on the intersection area of the target and swapped images. 
This metric helps to estimate the part of \textit{appearance leakage}.

\begin{table}
\begin{center}
\begin{adjustbox}{width=\columnwidth}

  \begin{tabular}{|l|l|l|l|}
    \hline
  
    Factor & Model & Criterion & Embedding  \\
    \hline
    ID sim &  ArcFace~\cite{arcface}  & cossim &  512-dimensinal parameter   \\
    HP dis & DECA~\cite{deca} & $L1$ &   euler angles    \\
    Exp dis & DECA~\cite{deca}  & $L1$ &  50-dimensional blendshape parameter      \\

    EG dis & RT-GENE~\cite{rtgene} & $L1$ & yaw and pitch        \\

    \hline
  \end{tabular}
\end{adjustbox}
\end{center}
\caption{\textbf{Specification of StyleGAN's $\mathcal{F/W+}$ quantitative analysis}. The cossim denotes the cosine similarity.}
\label{supp_table:analysis_spec}

\end{table}

\begin{figure}[t!]
    \centering 
    \includegraphics[width=\linewidth]{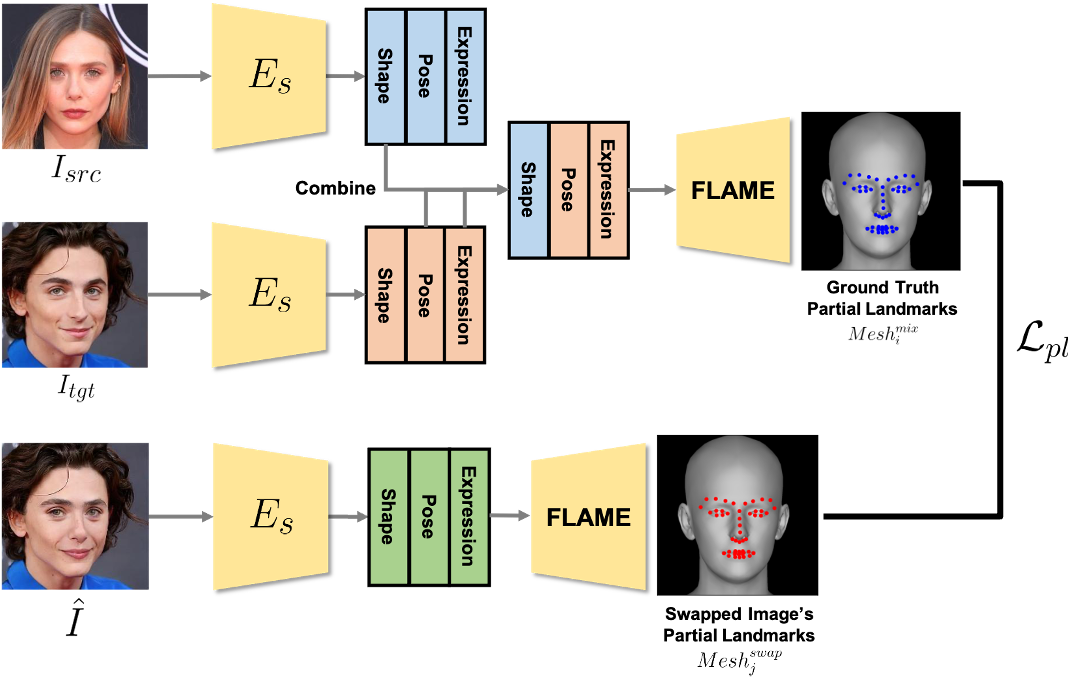}
    \caption{\textbf{Details of Partial Landmark Loss.} We designate the 51 inner facial feature related landmarks from 5023 vertices as the interest of loss. $Mesh_{i}^{mix}$ works as the ground truth of the partial landmarks which supervises the location of $Mesh_{j}^{swap}$.} 
    \label{fig:plloss}
\end{figure}

\section*{D. Details of Objective Functions}
\subsection*{D.1. Partial Landmark Loss.}
We proposed a novel partial landmark loss $L_{pl}$, which helps the swapped image's inner facial shape to resemble the source's one and follow the target's head pose and expression. Please see Fig.~\ref{fig:plloss}.
\subsection*{D.2. Adversarial Loss}
We follow the StyleGAN2~\cite{sg2}'s non-saturating adversarial loss and R1 regularizer. 
\begin{equation}    
    \mathcal{L}_{adv-D} = \text{softplus}(-D(I_{tgt})) + \text{softplus}(D(\hat{I})),
\end{equation}   
\begin{equation}    
    \mathcal{L}_{adv-G} = \text{softplus}(-D(\hat{I})),
\end{equation}   
\begin{equation}
    \mathcal{L}_{R1} = \frac{\gamma}{2} [\lVert \nabla_{\textbf{x}} D(\textbf{x}) \rVert^2_2],
\end{equation}   
\begin{equation}
    \mathcal{L}_{adv} = \mathcal{L}_{adv-D}+\mathcal{L}_{adv-G}+\mathcal{L}_{R1},
\end{equation}  
where $D$ is a pre-trained discriminator of StyleGAN2.

\section*{E. Comparison with StyleSwap}

\begin{figure*}[t!]
    \centering 
    \includegraphics[width=\linewidth]{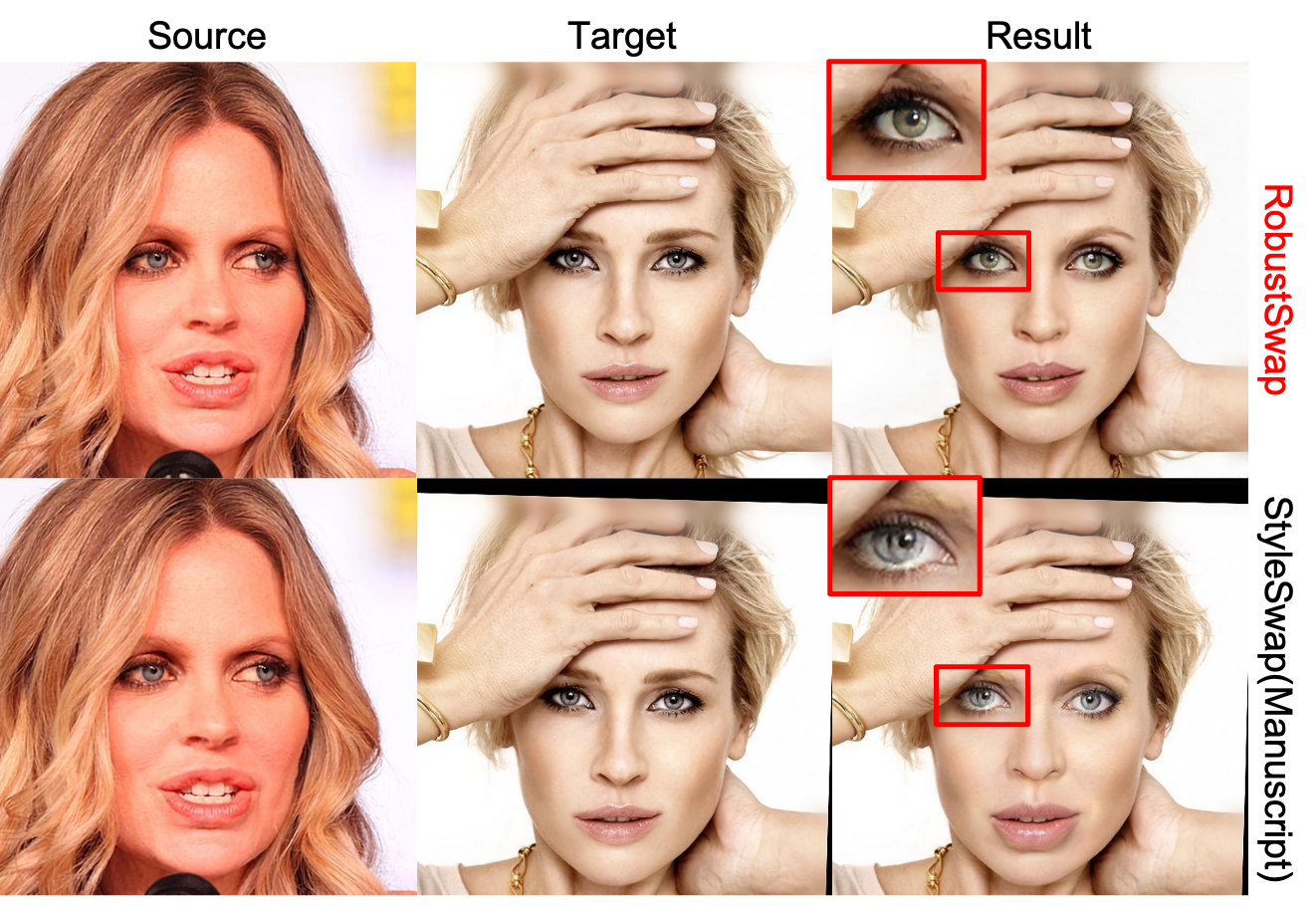}
    \caption{\textbf{Comparison with StyleSwap's Fig. 5 in the main manuscript.} Please pay attention to the \textcolor{red}{red} box which indicates the result's pupil. The source and target images are from CelebA-HQ train dataset.}
    \label{fig:retrieve}
\end{figure*}

\begin{figure*}[t!]
    \centering 
    \includegraphics[width=\linewidth]{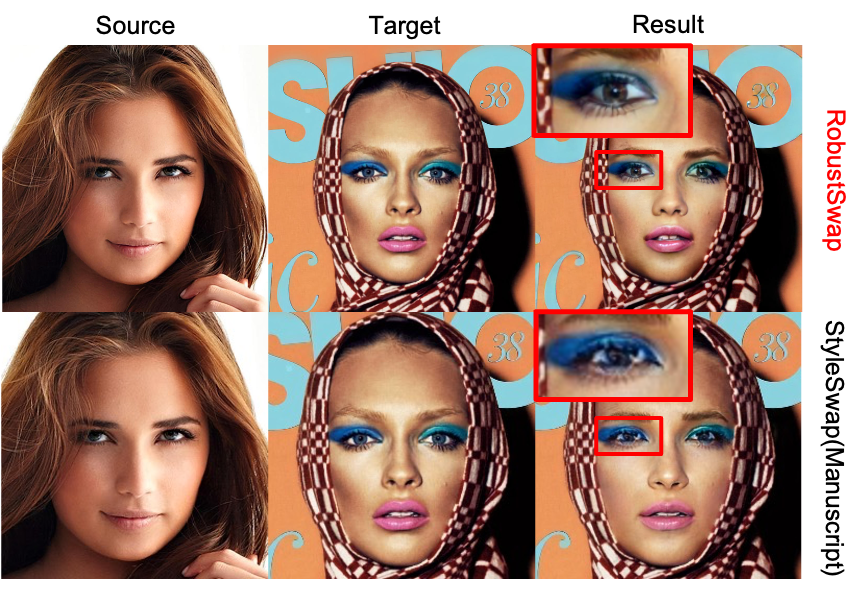}
    \caption{\textbf{Comparison with StyleSwap's Fig. 6 in the main manuscript.} Please pay attention to the \textcolor{red}{red} box which indicates the result's pupil. The source and target images are from CelebA-HQ train dataset.}
    \label{fig:retrieve2}
\end{figure*}

StyleSwap~\cite{styleswap} is also a state-of-the-art baseline that proposes a modified StyleGAN~\cite{sg2}-based architecture with the ToMask branch similar to ToRGB branch of the original StyleGAN. Although their open-source code is not released to the public, to prove the superiority of our \textbf{RobustSwap}, we retrieve the StyleSwap's source and target images and compare with \textbf{RobustSwap}'s result.
As can be seen in Fig.~\ref{fig:retrieve} and~\ref{fig:retrieve2}, StyleSwap fails to synthesize the pupil of the swapped image, while \textbf{RobustSwap} seamlessly reconstructs.

\begin{figure*}[t!]
    \centering 
    \includegraphics[width=\linewidth]{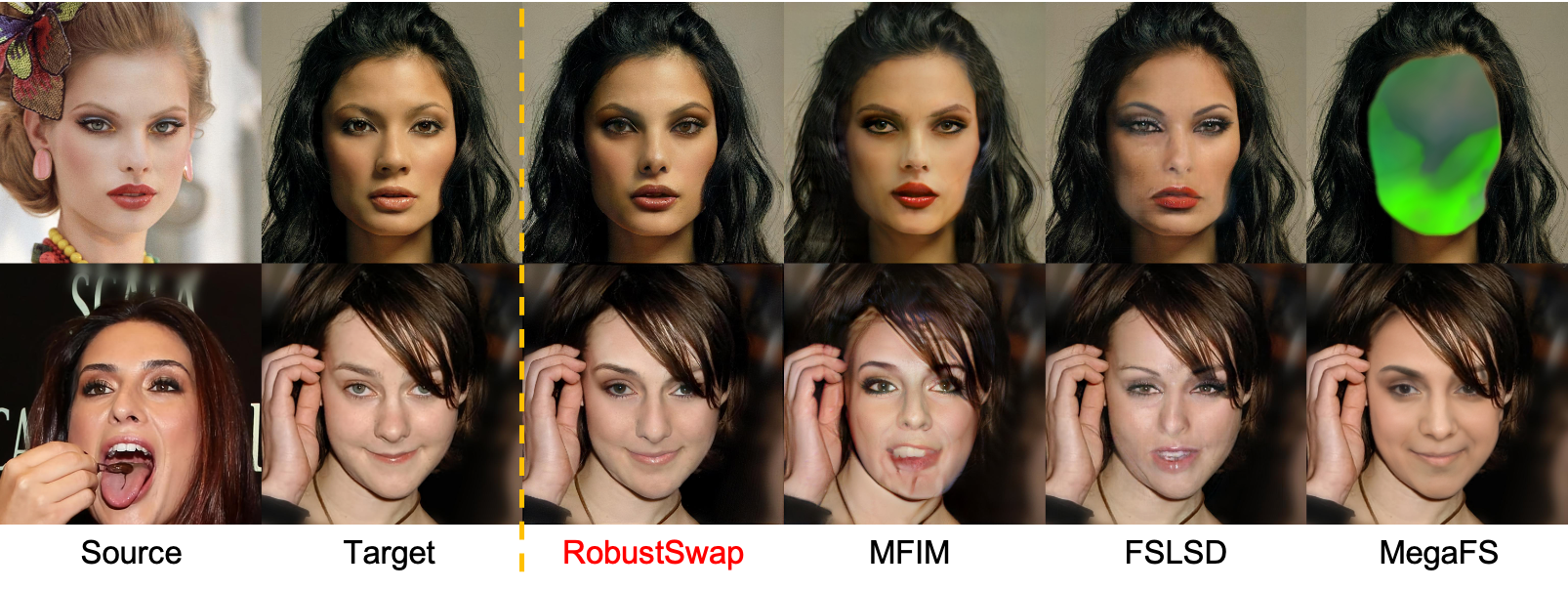}
    \caption{\textbf{Qualitative comparisons \#1} on 1024 × 1024 resolution same gender (female) CelebA-HQ with megapixel baselines}
    \label{fig:mega1}
\end{figure*}
\begin{figure*}[t!]
    \centering 
    \includegraphics[width=\linewidth]{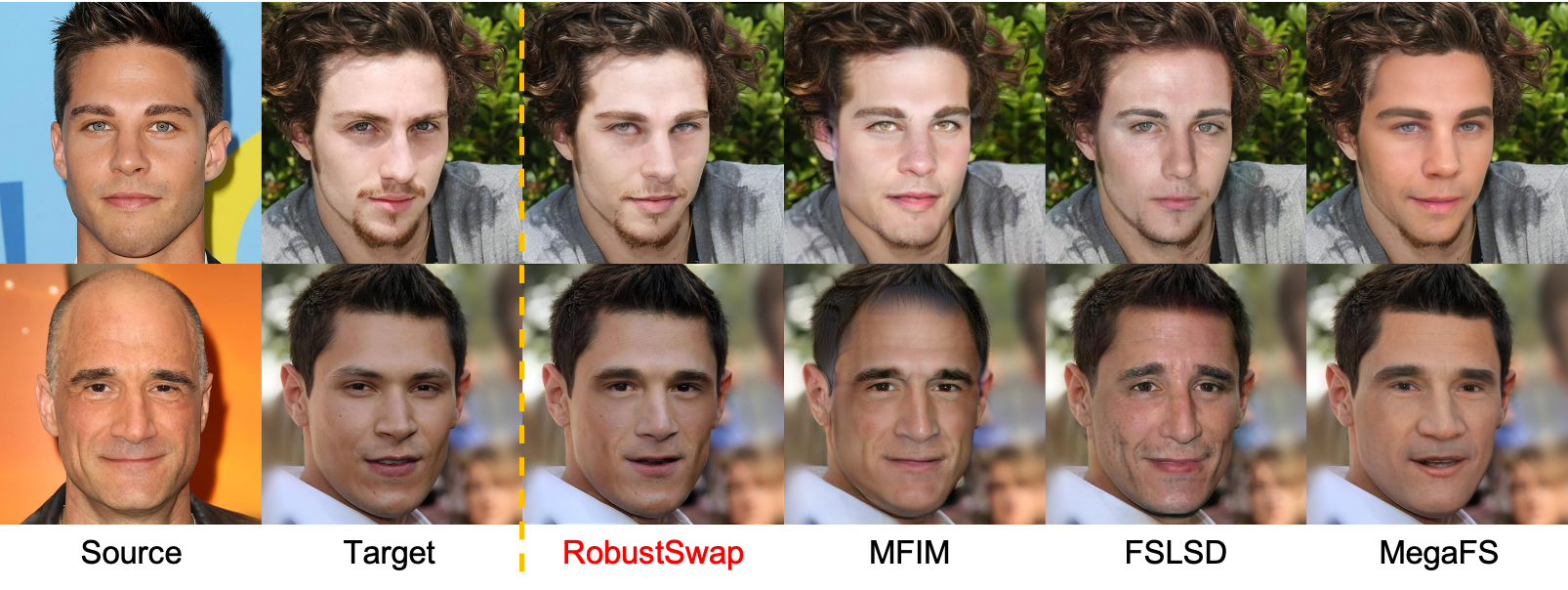}
    \caption{\textbf{Qualitative comparisons \#2} on 1024 × 1024 resolution same gender (male) CelebA-HQ with megapixel baselines}
    \label{fig:mega2}
\end{figure*}
\begin{figure*}[t!]
    \centering 
    \includegraphics[width=\linewidth]{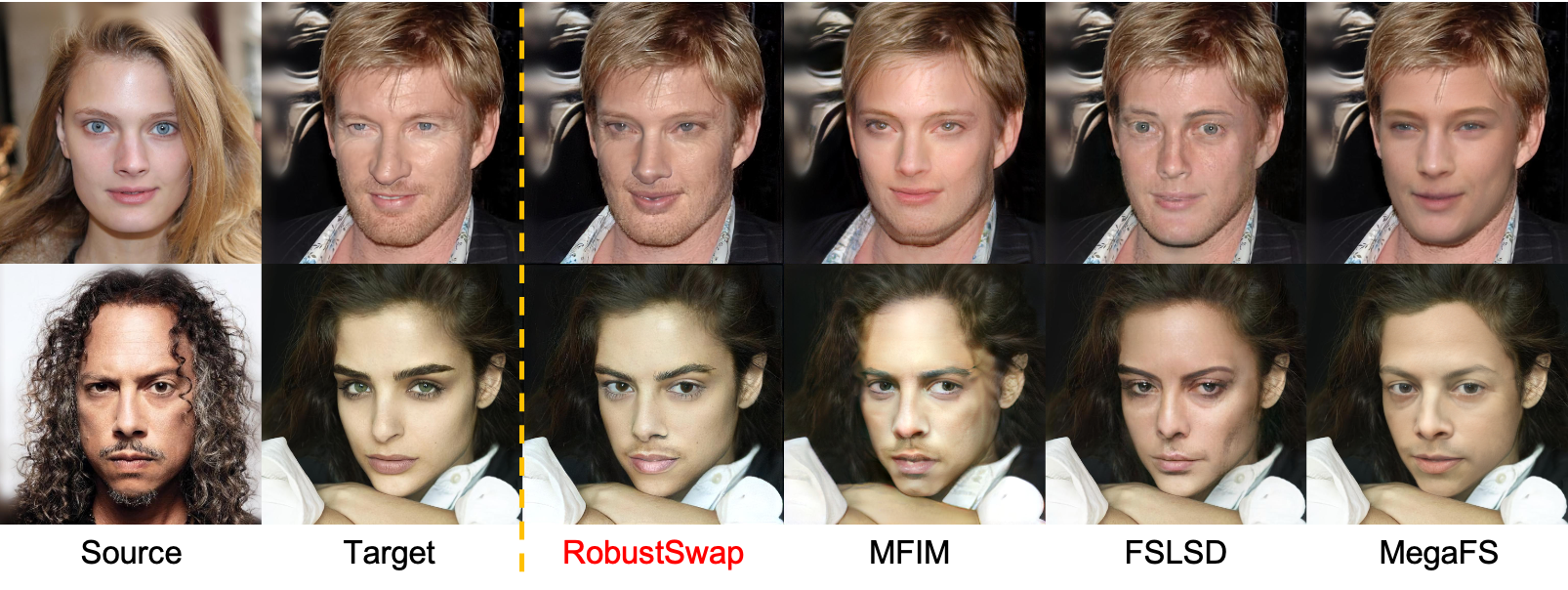}
    \caption{\textbf{Qualitative comparisons \#3} on 1024 × 1024 resolution cross gender CelebA-HQ with megapixel baselines}
    \label{fig:mega3}
\end{figure*}

\begin{figure*}[t!]
    \centering 
    \includegraphics[width=\linewidth]{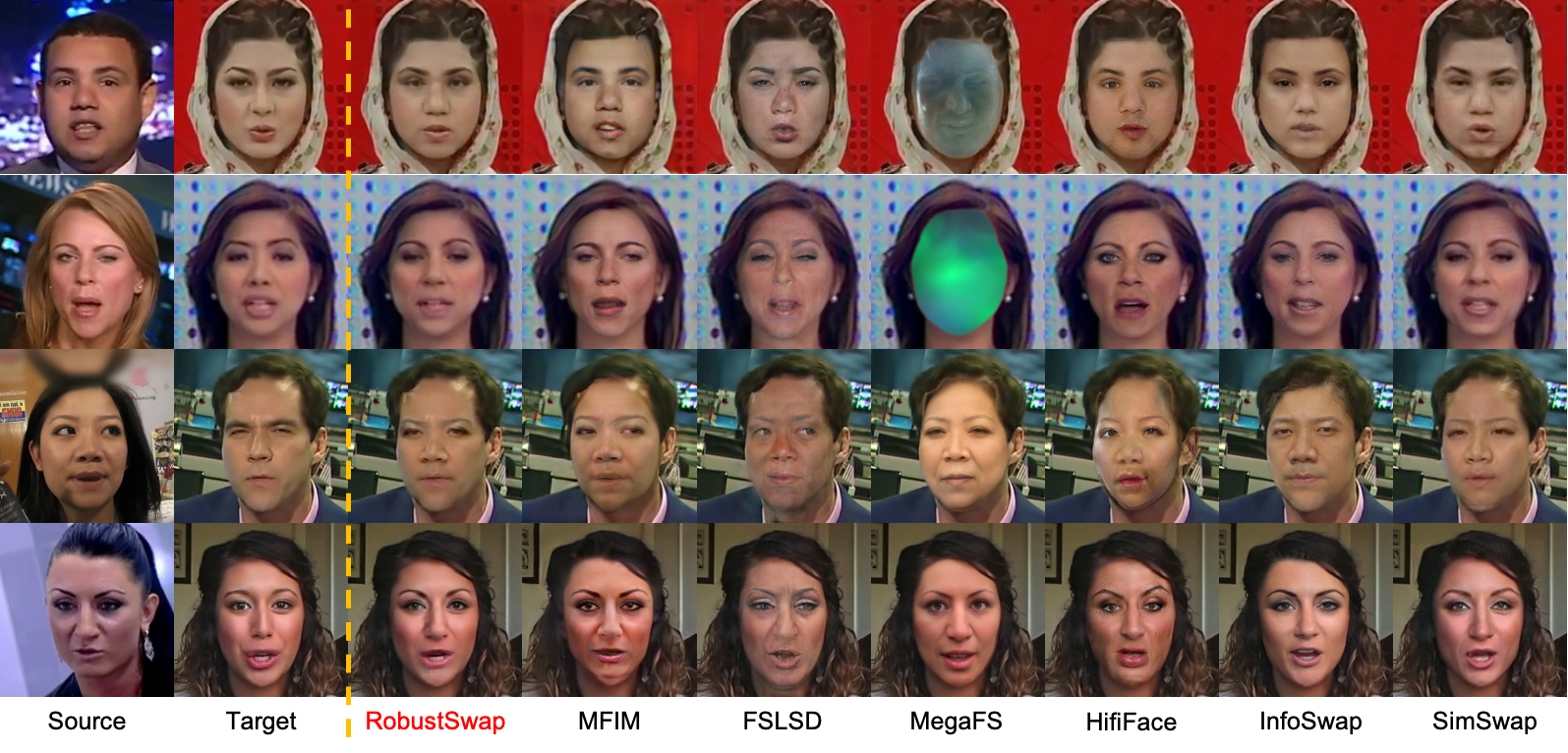}
    \caption{\textbf{Qualitative comparisons \#4} on 256 × 256 resolution FF++ with all baseline}
    \label{fig:FF++1}
\end{figure*}
\begin{figure*}[t!]
    \centering 
    \includegraphics[width=\linewidth]{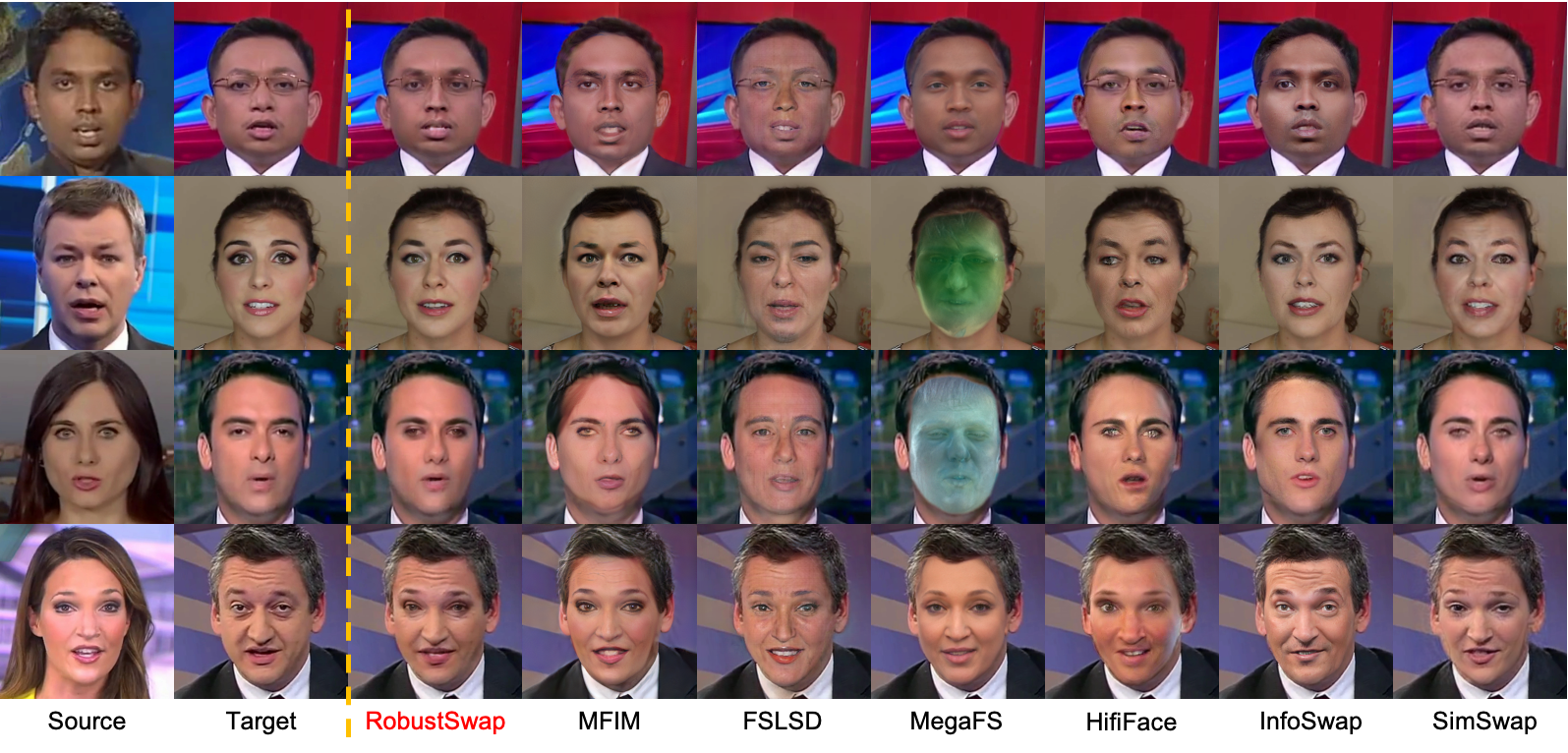}
    \caption{\textbf{Qualitative comparisons \#5} on 256 × 256 resolution FF++ with all baseline}
    \label{fig:FF++2}
\end{figure*}

\section*{F. More Comparisons}
As mentioned before, we show more numerous results as extension of Fig. 7 and 8 in the main manuscript, comparison with all baselines and megapixel baselines, respectively (from Fig~\ref{fig:mega1} to~\ref{fig:FF++2}). Note that the video comparisons are in the attached .mp4 file, please watch the video.

\begin{figure*}[t!]
    \centering 
    \includegraphics[width=\linewidth]{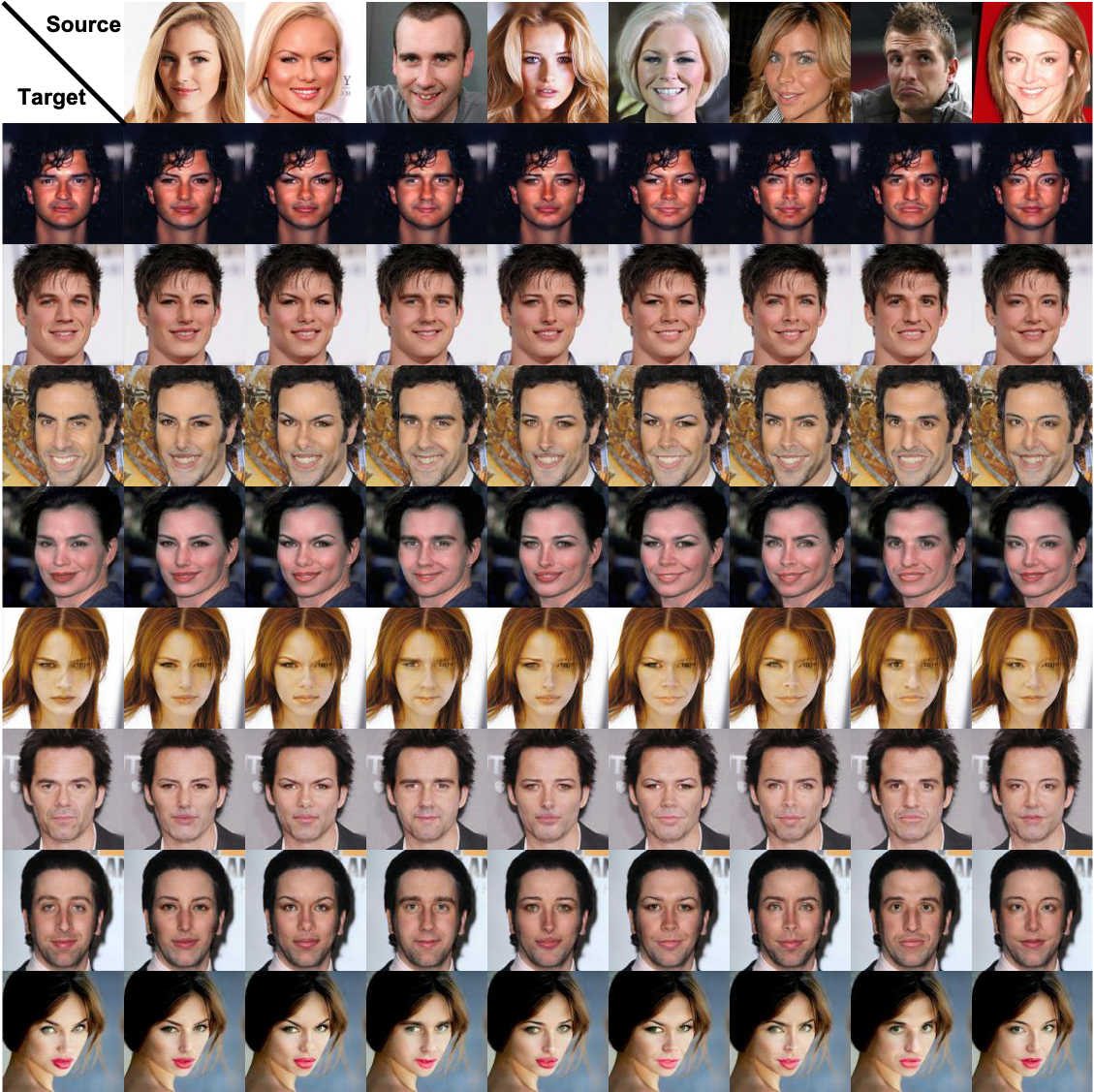}
    \caption{\textbf{Face matrix \#1}  on \textbf{RobustSwap} from 1024 x 1024 CelebA-HQ}
    \label{fig:celeba_matrix1}
\end{figure*}

\begin{figure*}[t!]
    \centering 
    \includegraphics[width=\linewidth]{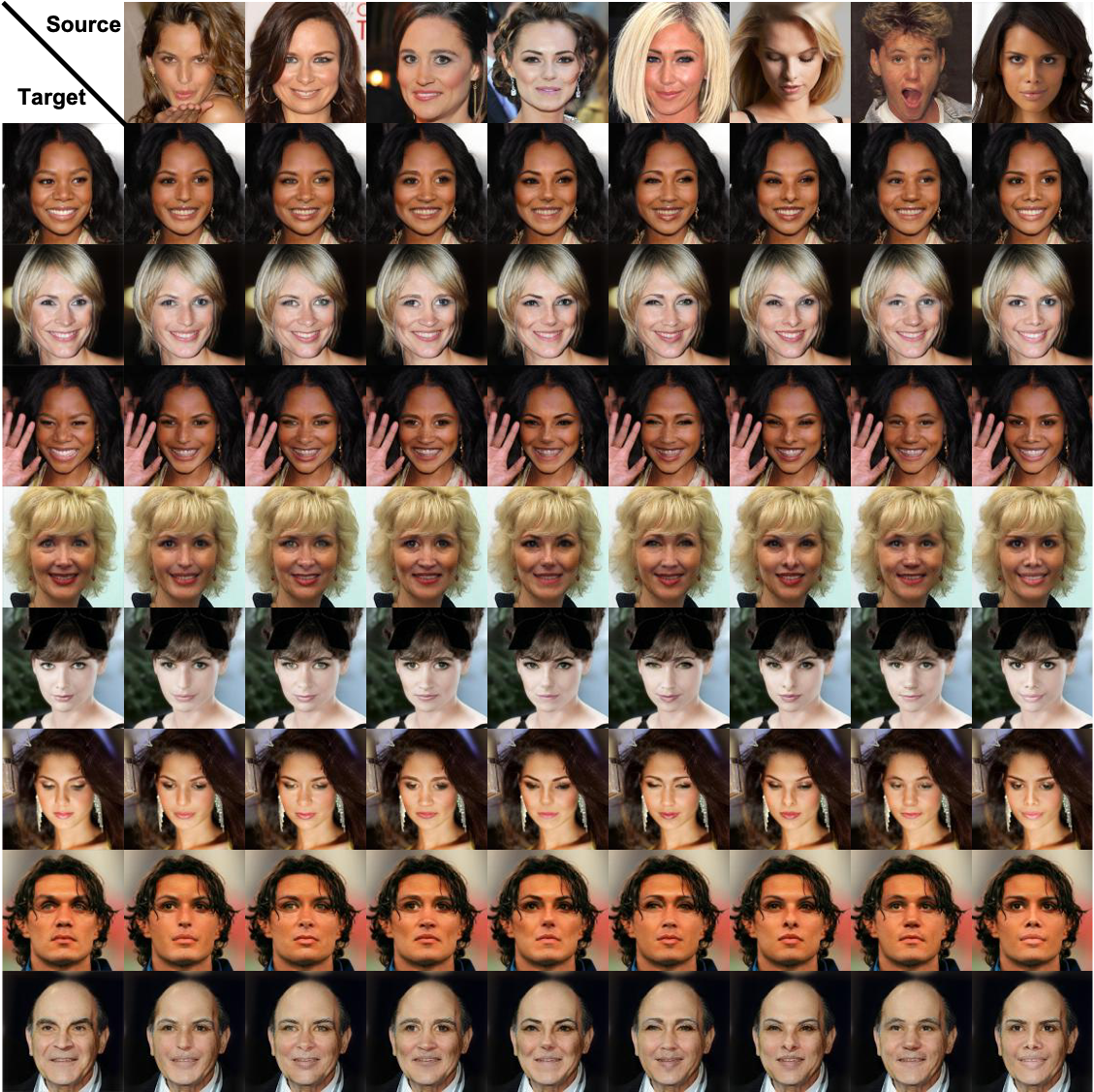}
    \caption{\textbf{Face matrix \#2} on \textbf{RobustSwap} from 1024 x 1024 CelebA-HQ}
    \label{fig:celeba_matrix2}
\end{figure*}

\begin{figure*}[t!]
    \centering 
    \includegraphics[width=\linewidth]{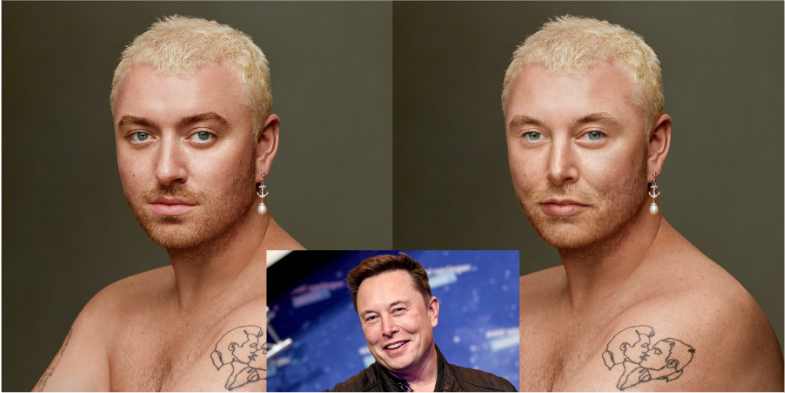}
    \caption{\textbf{In-the-wild result.} Source: Elon Musk \& Target: Sam Smith}
    \label{fig:inthewild1}
\end{figure*}

\begin{figure*}[t!]
    \centering 
    \includegraphics[width=\linewidth]{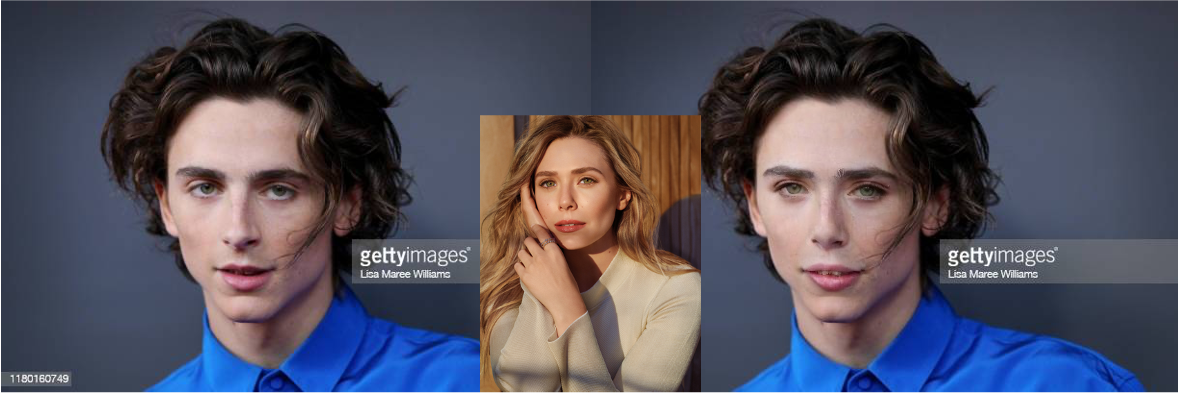}
    \caption{\textbf{In-the-wild result.} Source: Elizabeth Olsen \& Target: Timothee Chalamet}
    \label{fig:inthewild2}
\end{figure*}
\begin{figure*}[t!]
    \centering 
    \includegraphics[width=\linewidth]{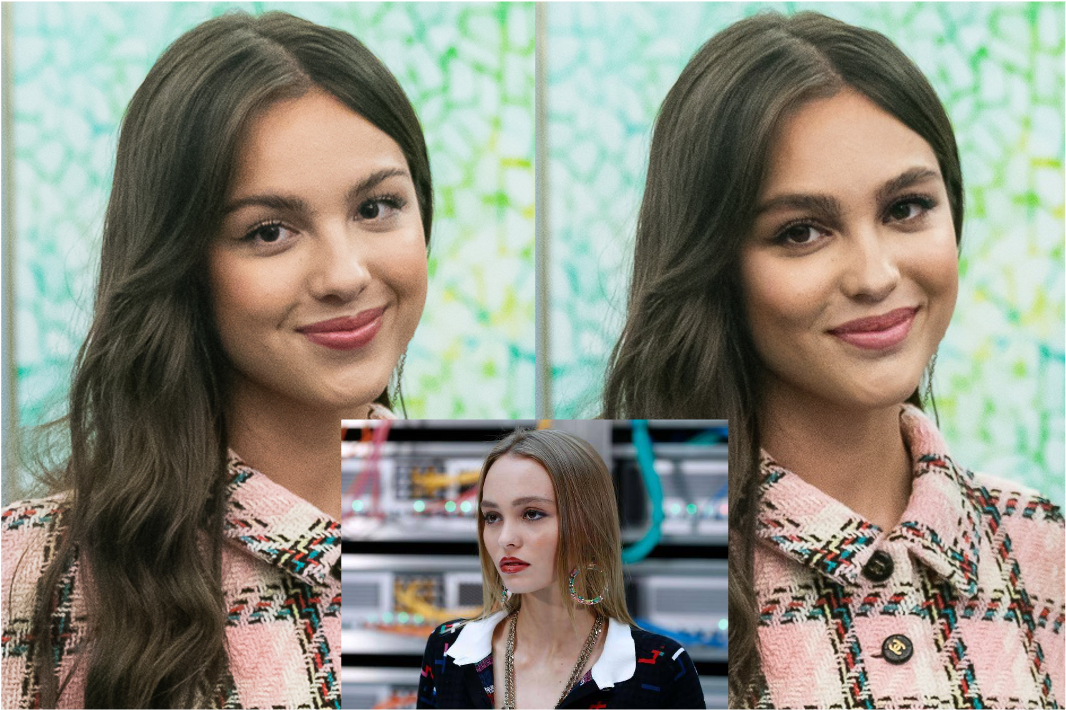}
    \caption{\textbf{In-the-wild result.} Source: Lily-Rose Depp \& Target: Olivia Rodrigo}
    \label{fig:inthewild3}
\end{figure*}
\begin{figure*}[t!]
    \centering 
    \includegraphics[width=\linewidth]{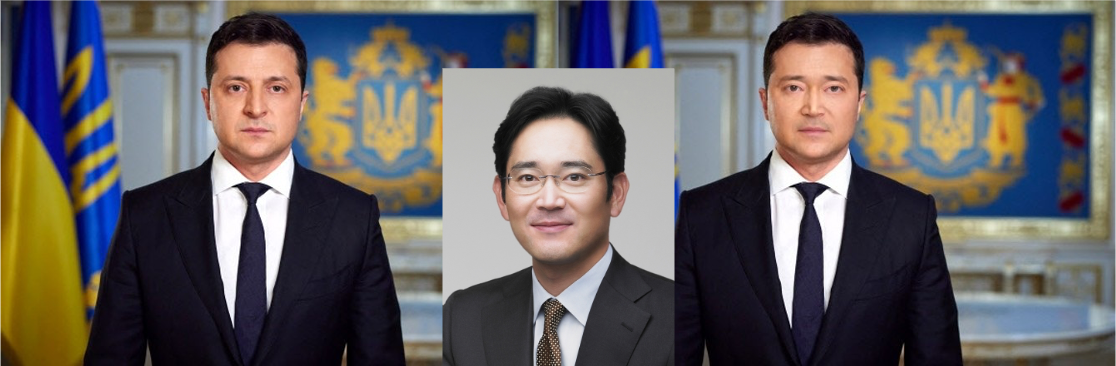}
    \caption{\textbf{In-the-wild result.} Source: Jaeyong Lee \& Target: Volodymyr Zelenskyy}
    \label{fig:inthewild4}
\end{figure*}

\section*{G. More Results}
From Fig.~\ref{fig:celeba_matrix1} to~\ref{fig:inthewild4}, we show the face matrices from FaceForensic++ (FF++)~\cite{ff++} dataset and internet-crawled in-the-wild data. Note that the video results are in the attached mp4 file. Please watch the video.
\end{document}